\documentclass[%
reprint,
superscriptaddress,https://www.overleaf.com/project/60418f9384090c596f682ce0
 nofootinbib,
 amsmath,amssymb,
 aps,
 pre,
]{revtex4-2}

\usepackage{graphicx}
\usepackage{dcolumn}
\usepackage{bm}
\usepackage{bbm}
\usepackage{float}
\usepackage{subcaption}
\usepackage{multirow}
\usepackage{comment}
\usepackage{dsfont}
\usepackage{amsthm}
\usepackage{amsmath}
\usepackage{physics}
\usepackage{xcolor}
\usepackage{makecell}
\usepackage{scalerel,stackengine}

\def\ie{{\frenchspacing\it i.e.}}
\def\eg{{\frenchspacing\it e.g.}}

\newcommand{\mat}[1]{\mathbf{#1}}

\def\f{\mat{f}}

\def\A{{\bf A}}
\def\B{{\bf B}}

\def\G{{\bf G}}

\def\x{\textbf{x}}

\def\z{\textbf{z}}

\newcommand{\red}[1]{{\color{black!30!red} #1}}
\newcommand{\green}[1]{{\color{black!50!green} #1}}

\def\spose#1{\hbox to 0pt{#1\hss}}
\def\simlt{\mathrel{\spose{\lower 3pt\hbox{$\mathchar"218$}}
     \raise 2.0pt\hbox{$\mathchar"13C$}}}
\def\simgt{\mathrel{\spose{\lower 3pt\hbox{$\mathchar"218$}}
     \raise 2.0pt\hbox{$\mathchar"13E$}}}
\def\simpropto{\mathrel{\spose{\lower 3pt\hbox{$\mathchar"218$}}
     \raise 2.0pt\hbox{$\propto$}}}

\def\beq#1{\begin{equation}\label{#1}}
\def\eeq{\end{equation}}
\def\beqa#1{\begin{eqnarray}\label{#1}}
\def\eeqa{\end{eqnarray}}
\def\eq#1{equation~(\ref{#1})}

\newtheorem{definition}{Definition}[section]

\begin{document}


\title{AI Poincar\'{e} 2.0: Machine Learning Conservation Laws from Differential Equations}

\author{Ziming Liu}
\affiliation{
Department of Physics, Massachusetts Institute of Technology, Cambridge, USA}
\author{Varun Madhavan}
\affiliation{
Indian Institute of Technology Kharagpur, India}
\author{Max Tegmark}
\affiliation{
Department of Physics, Massachusetts Institute of Technology, Cambridge, USA}

\date{\today}

\begin{abstract}
We present a machine learning algorithm that discovers conservation laws from differential equations, both 
numerically (parametrized as neural networks) and symbolically,
ensuring their functional independence (a non-linear generalization of linear independence). 
Our independence module can be viewed as a nonlinear generalization of singular value decomposition. 
Our method can readily handle inductive biases for conservation laws. We validate it with examples
including the 3-body problem, the KdV equation and nonlinear Schr\"odinger equation. 
\end{abstract}

\maketitle

\section{Introduction}

\begin{figure}
    \centering
    \includegraphics[width=1.0\linewidth]{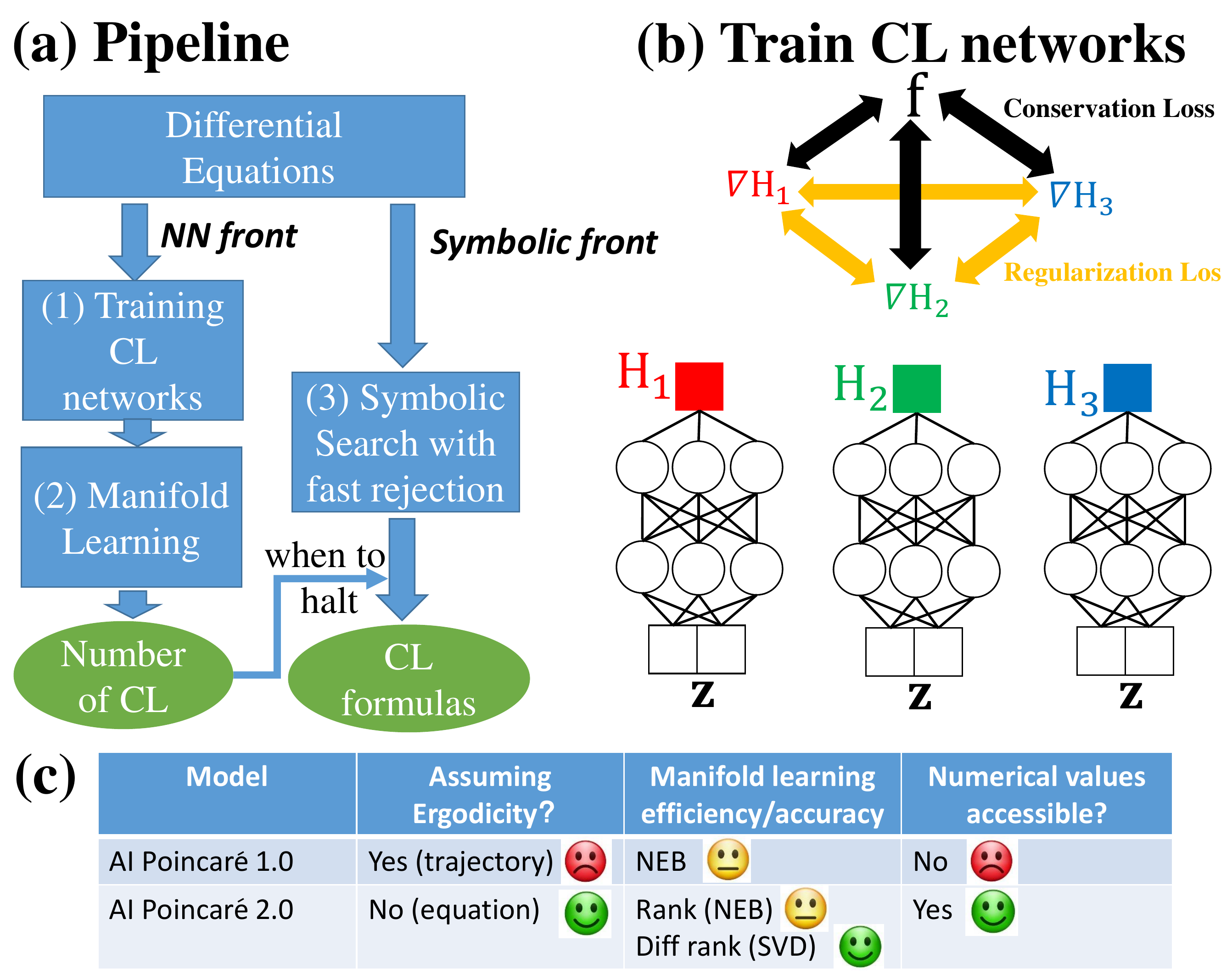}
    \caption{(a) The AI Poincar\'{e} 2.0 pipeline: The NN front leverages neural networks for conservation laws, while the symbolic front searches for formulas with fast rejection. (b) Training is minimizing each network's conservation loss combined with a function dependence penalty. (c) Comparing 1.0 and 2.0. NEB refers to Neural Empirical Bayes, the manifold learning algorithm we adopted in 1.0.}
    \label{fig:model}
    \vskip -0.5cm
\end{figure}

The importance of conservation laws (CLs) in physics can hardly be overstated~\cite{PhilipAnderson1972}. Physicists usually derive conservation laws with time-consuming pencil and paper methods, using different hand-crafted strategies for each specific problem. This motivates searching for a general-purpose problem-agnostic approach. A few recent papers have exploited machine learning to auto-discover conservation laws~\cite{poincare,mototake2019interpretable, PhysRevResearch.2.033499,ha2021discovering}. Despite promising preliminary results, these techniques are not guaranteed to discover {\it all} conservation laws. In this paper, we start with differential equations defining a dynamical system and 
aim to discover all its conservations laws, either in numerical form (parameterized as neural networks) or in symbolic form. 
The new method is named AI Poincar\'{e} 2.0 since it builds on ~\cite{poincare}. When no confusion occurs, we call the original method 1.0, and the new method 2.0. We summarize three major improvements of 2.0 over 1.0 below, as well as in FIG.~ \ref{fig:model}(c).

First, 1.0 tacitly requires the assumption that the trajectory is ergodic, while 2.0 does not need the assumption since it directly deals with differential equations. 2.0 can apply to systems with dissipation or directionality on which 1.0 falls short. A case of directionality is the Korteweg–De Vries (KdV) wave equation, where solitons travel from left to right, violating ergodicity.

Second, 2.0 introduces a new manifold learning method that is more efficient and accurate than 1.0. 2.0 also extends the notion of variable dependence to functional dependence, which is fundamental and useful for physics and machine learning applications.

Third, 2.0 provides numerical evaluation of each conserved quantity, while 1.0 provides no information at all other than the conserved quantity exists. These numerical values can hopefully give physicists insights about properties or symbolic forms of the conservation laws.



\begin{figure*}
    \centering
    \includegraphics[width=1\linewidth]{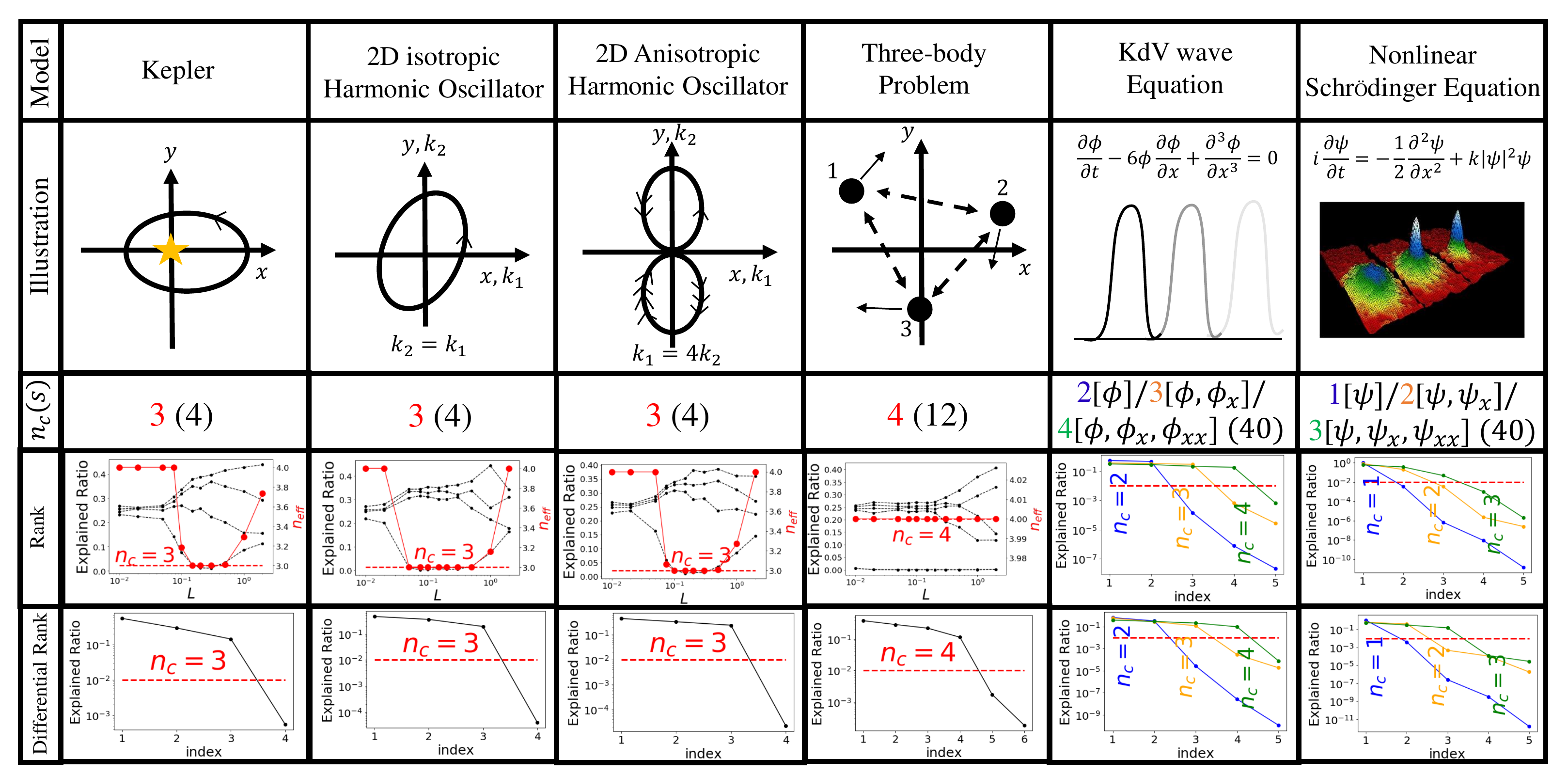}
    \caption{Tested ordinary and partial differential equation examples, each of which has $s$ degrees of freedom and $n_c$ conservation laws. AI Poincar{\'e} 2.0 is seen to find the correct $n_c$ by computing rank (read off as the low flat region of the $n_{\rm eff}$ curve as defined in \cite{poincare})
    or differential rank.}
\label{fig:examples}
\vskip -0.0cm
\end{figure*}

In the Method section, we introduce our notation and the AI Poincar\'{e} 2.0 algorithm. In the Results section, we apply AI Poincar\'{e} 2.0 to various systems (illustrated in FIG.~\ref{fig:examples}) to test its ability to auto-discover conservation laws, followed by discussions and conclusions. We note other works exploring the direction of ``machine learning meets conservation laws"~\cite{wang2019learning,sturm2022conservation,kunin2020neural}, which have different goals than ours.

\section{Method}
\subsection{Problem and Notation} We consider a first-order ordinary differential equation (ODE) $\frac{d\z}{dt}=\f(\z)$ where $\z\in\mathbb{R}^s$ is the state vector and $\f:\mathbb{R}^s\to\mathbb{R}^s$ is a vector field. {\it Hamiltonian systems} correspond to the special case where $s$ is even and $\f=\left(\frac{\partial{H_0}}{\partial \mat{p}},-\frac{\partial H_0}{\partial \x}\right)$ for a Hamiltonian function $H_0$. A {\it conserved quantity} is a scalar function $H(\z)$ whose value remains constant along a trajectory $\z(t)$ determined by $\frac{d\z}{dt}=\f(\z)$ with any initial condition $\z(t=0)=\z_0$. A necessary and sufficient condition for a scalar function $H(\z)$ being a conservation law is $\nabla H\cdot \f= 0$, because $\frac{d}{dt}H\left(\z(t)\right)=\nabla H\cdot \frac{d\z}{dt}= \nabla H\cdot\f$. We use hats to denote unit vectors, e.g., $\widehat{\f}\equiv\f/|\f|$. Our goal is to discover the maximal number $n_c$ independent conserved quantities $\{H_1(\z),H_2(\z),\cdots,H_{n_c}(\z)\}$ numerically and symbolically, optionally with user-specified properties.

Dynamical systems of the form $\frac{d\z}{dt}=\f(\z)$ are very general because (1) higher-order ODEs, e.g. Newtonian mechanics, can always be transformed to first-order ODEs by including derivatives as new variables in $\z$, and (2) partial differential equations (PDEs) can be approximated by ODEs by discretizing space.

\subsection{AI Poincar{\'e} 2.0} 
AI Poincar{\'e} 2.0 consists of three steps: (1) learn conservation laws parameterized by neural networks, (2) count the number of independent conservation laws and (3) find symbolic formulas for conservation laws. The pipeline is illustrated in FIG.~\ref{fig:model}.

\subsubsection{ Parameterizing conservation laws by neural networks}

We parameterize a conserved quantity as a neural network $H(\z;{\boldsymbol \theta})$ where ${\boldsymbol \theta}$ are model parameters. Our loss function is defined as
\begin{equation}
    \ell(\theta) \equiv \frac{1}{P}\sum_{i=1}^{P} \left| \widehat{\f}(\z^{(i)})\cdot \widehat{\nabla H} (\z^{(i)};\boldsymbol\theta)\right|^2,
\end{equation}
where $\z^{(i)}$ denotes the $i^{\rm th}$ sample in phase space. $\nabla H(\z)$ can be easily computed with automatic differentiation~\cite{goodfellow2016deep}. Note that $\widehat{\f}$ and $\widehat{\nabla H}$ are normalized unit vectors, to make the loss function dimensionless and invariant under uninteresting re-scaling of $H$. We update $\boldsymbol\theta$ by trying to minimize the loss function until it drops below a small threshold $\epsilon$.

To obtain multiple conserved quantities, one can repeat the above method with different random seeds and hope to discover algebraically independent ones. In practice, however, we find that learned conservation laws are often highly correlated for different initializations~\footnote{This seems to imply some `simpler' conservation laws are preferred by neural networks over others.}. To encourage linear independence between two neural networks, say, $H_1$ and $H_2$, we add a regularization term
\begin{equation}
    R(\boldsymbol\theta_1,\boldsymbol\theta_2) \equiv \frac{1}{P}\sum_{i=1}^P \left|  \widehat{\nabla H_1} (\z^{(i)};\boldsymbol\theta_1)\cdot\widehat{\nabla H_2} (\z^{(i)};\boldsymbol\theta_2)\right|^2
\end{equation}
to the loss function. 
Since we know that there cannot be more conservation laws than degrees of freedom $s$, we train $n = s$ models together by minimizing the loss function $\ell_1+\lambda\ell_2$ defined by
\begin{equation}\label{eq:loss}
    \ell = \underbrace{\frac{1}{n}\sum_{i=1}^n \ell(\boldsymbol\theta_i)}_{\ell_1}  + \lambda\times \underbrace{ \frac{2}{n(n-1)}\sum_{i=1}^n\sum_{j=i+1}^n R(\boldsymbol\theta_i,\boldsymbol\theta_j)}_{\ell_2},
\end{equation}
where $\lambda$ is a penalty coefficient. We refer to $\ell_1$ and $\ell_2$ as {\it conservation loss} and {\it independence loss}, respectively.

\subsubsection{Counting the number of independent conserved quantities}
After training, we aim to determine (in)dependence among these neural networks. 
Specifically, we are interested in  {\it functional independence}, a direct generalization of linear independence 
that we define and compute as described below.

\begin{definition}
{\rm \bf Functional independence}.
A set of non-zero functions $H_1(\z)$, $H_2(\z),\cdots,H_n(\z)$ is independent if 
\beq{FunctionalIndepEq}
    f(H_1(\z),H_2(\z),\cdots,H_n(\z)) = 0 \Longrightarrow f=0
\eeq
or, equivalently, if no function $H_i(\z)$ can be constructed from (possibly nonlinear and multivalued) combinations of the other functions. 
\end{definition}
\begin{definition}
{\rm \bf Function set rank}.
The function set $\mathcal{H}=\{H_1(\z), H_2(\z), \cdots H_n(\z)\}$ has rank $k\leq n$ if it contains $k$ but not $k+1$ functions that are independent.
\end{definition}

{\bf Computing the function set rank} We determine the rank $k$ with a nonlinear manifold learning method. We
define the matrix $\A$ such that $A_{ij}$ is the value of the $j^{\rm th}$ neural network evaluated at the $i^{\rm th}$ sample point:
\begin{equation}
\A = 
\begin{pmatrix}
H_1(\z^{(1)}) & H_2(\z^{(1)}) & \cdots & H_n(\z^{(1)}) \\
    H_1(\z^{(2)}) & H_2(\z^{(2)}) & \cdots & H_n(\z^{(2)}) \\
    \cdots & \cdots & \cdots & \cdots \\
    H_1(\z^{(P)}) &
    H_2(\z^{(P)}) & \cdots & H_n(\z^{(P)}) \\
\end{pmatrix},
\end{equation}
where $P\gg n$ is the number of data points $\z^{(i)}$. 
If we interpret each row of $\A$ as a point in $\mathbb{R}^n$, then the matrix corresponds to a point cloud in $\mathbb{R}^n$ located on a a manifold, whose dimensionality $k$ is equal to the function set rank. 
If there are $k$ independent {\it linear} conserved quantities (where $H_i(\z)$ are linear functions),
then the point cloud will lie on a $k$-dimensional hyperplane that can readily be discovered using singular value decomposition (SVD):
$k$ is then the number of non-zero singular values, \ie, the rank of the matrix $\A$.
For our more general nonlinear case, we wish to discover the manifold that the point cloud lies on even if it is curved.
For this, we exploit the manifold learning algorithm proposed in Poincar\'{e} 1.0~\cite{poincare} to measure the manifold dimensionality~\footnote{Although the nonlinear manifold learning method introduced in AI Poincar{\'e} 1.0 also applies here, the ways to compute the number of conserved quantities $n_c$ is different and actually \textit{dual}. In Poincar{\'e} 1.0, $n_c$ is the phase space dimension minus the dimension of the trajectory manifold. While in this paper, $n_c$ is equal to the dimension of the manifold. Because of this duality, the explained ratio diagram (ERD) in Poincar\'{e} 1.0 resembles a hill while in FIG.~\ref{fig:examples} the ERD is upside down and resembles a valley.}, which performs local Monte Carlo sampling followed by a linear dimensionality estimation method, from which we define $n_{\rm eff}$. For the rank row in FIG.~\ref{fig:examples} (excluding the two last PDE examples), $n_c$ can be readily read off as the value of $n_{\rm eff}$ corresponding to the low flat valley.

Taking the derivative of $f(H_1(\z),H_2(\z)\cdots, H_n(\z))=0$ from \eq{FunctionalIndepEq}
with respect to $z_i$ gives.
\begin{equation}
\label{eq:B}
\begin{aligned}
    \underbrace{\begin{pmatrix}
    H_{1,1} & H_{2,1} & \cdots & H_{n,1} \\
    H_{1,2} & H_{2,2} & \cdots & H_{n,2} \\
    \vdots & \vdots &  & \vdots \\
    H_{1,s} & H_{2,s} & \cdots & H_{n,s}
    \end{pmatrix}}_{\mat{B}}
    \underbrace{\begin{pmatrix}
    f_{,1} \\
    f_{,2} \\
    \vdots \\
    f_{,n}
    \end{pmatrix}}_{\mat{\nabla f}}=\mat{0}.
\end{aligned}
\end{equation}
This means that, if $\{H_1,\cdots,H_n\}$ and $f$ are differentiable functions and $\B$ has full rank, then
$\nabla f(\z)$ and therefore $f(z)$ itself must vanish identically, so the functions $H_i$ must be independent.
We exploit this to define {\it differentiable independence} and {\it differentiable rank} as follows:
\begin{definition}
{\rm \bf Differential functional independence}.
A set of $n$ non-zero differentiable functions $\mathcal{H}$ is differentially independent 
if their gradients are linearly independent, \ie, if
${\rm rank}\ \mat{B}(\z) = n$
almost everywhere (for all $\z$ except for a set of measure zero).
\end{definition}

\begin{definition}
{\rm \bf differential function set rank}.
The differential rank of the function set $\mathcal{H}=\{H_1(\z), H_2(\z), \cdots H_n(\z)\}$ is defined as $k_D=\underset{\z}{\rm max}\ {\rm rank}\ \mat{B}(\z)$.
\end{definition}
In practice, it suffices to compute the maximum over a finite number of points $P\gg n$:
it is exponentially unlikely that such sampling will underestimate the true manifold dimensionality,
just as it is  exponentially unlikely that 
$P$ random points in 3-dimensional space will happen to lie on a plane.

Numerically, one can apply singular value decomposition to $\mat{B}$ to obtain singular values $\{\sigma_1,\sigma_2,\cdots,\sigma_n\}$, and define the rank as the number of non-zero singular values. In practice, we treat components as vanishing if the explained 
fraction of the total variance,  $\sigma_i^2/\sum_j\sigma_j^2$, is below 
$\epsilon=10^{-2}$. In the differential rank row of FIG.~\ref{fig:examples} (plus two PDE examples in the rank row), we draw a horizontal line at $\epsilon$, and define $n_c$ as the number of components above that line. The differential rank and the rank mostly give consistent results, as shown in FIG.~\ref{fig:examples}. However, the differential rank is more efficient to compute and appears to be more stable in high dimensions (see examples in Section \ref{sec:three_body}).


\subsubsection{Discovering symbolic formulas}
When no domain knowledge is available for a physical system, we perform a brute-force search over symbolic formulas ordered by increasing complexity as in ~\cite{feynman1,feynman2}. We leverage the criterion $ \hat{\f}\cdot\widehat{\nabla H}=0$ to determine if a candidate function $H(\z)$ is a conserved quantity or not. We implement a brute force algorithm in C++ for speed and employ a fast rejection strategy for further speedup: we prepare $n_p=10$ test points in advance, and reject $H$ immediately if $\left| \widehat{\f}(\z)\cdot\widehat{\nabla H} (\z)\right|>\epsilon_s=10^{-4}$ for any test point $\z$. If a formula survives at the $n_p$ test points, we test thoroughly by checking the condition numerically on the whole dataset, or test the condition symbolically. We determine whether the new conserved quantity is independent of already discovered ones by checking if the differential function set rank increases by 1 when adding the new conserved quantity. Appendix \ref{app:symbolic_dependence} provides further technical details.

{\bf Including inductive biases to learn conservation laws}  
Above we did not distinguish between integrals of motion (IOM) and conservation laws. Loosely speaking, conservation laws are those IOMs with inductive biases. As clarified in ~\cite{landau1976mechanics} and Section \ref{app:integrability}, conservation laws are usually derived from homogeneity and isotropy of space and time, and have the feature of being additive, \ie, expressible as a sum of simple terms involving only a small subset of the degrees of freedom. 
Conserved quantities of PDEs usually take the form of integrals over space.
We incorporate any such desired inductive biases into our method by restricting the neural networks parametrizing $H_i(\z)$ to have the corresponding properties. 

\section{Results}
\begin{table*}[htbp]
	\centering
	\begin{tabular}{|c|c|c|c|}\hline
		System& Integrals of Motion or Conservation Laws & Reverse Polish Notation &  Discovered \\\hline
    	\multirow{3}{*}{Kepler Problem}& $H_1=\frac{1}{2}(p_x^2+p_y^2)-\frac{1}{\sqrt{x^2+y^2}}$ & \texttt{$p_x$Q$p_y$Q+rIo-} & Yes\\\cline{2-4}
		& $H_2=xp_y-yp_x$ & \texttt{x$p_y$*y$p_x$*-} &  Yes\\\cline{2-4}
		& $H_3=(xp_y-yp_x)p_y+\hat{r}_x$ & \texttt{x$p_y$*y$p_x$*-$p_y$*xr/+} & No\\\hline
		1D Damped Oscillator & $H_1 = {\rm arctan}(\frac{p}{x})+{\rm ln}\sqrt{x^2+p^2}/\gamma$ & \texttt{px/TxQpQ+RL$\gamma$/+} & No \\\hline
		\multirow{3}{*}{\makecell{Isotropic Oscillator}}& $H_1=\frac{1}{2}(x^2+p_x^2)$ & \texttt{xQ*$p_x$Q+} & Yes \\\cline{2-4}
		& $H_2=\frac{1}{2}(y^2+p_y^2)$ & \texttt{yQ$p_y$Q+} & Yes \\\cline{2-4}
		& $H_3=xy+p_xp_y$ & \texttt{xy*$p_xp_y$*+} & Yes \\\hline
		\multirow{3}{*}{\makecell{ Anisotropic Oscillator}}& $H_1=\frac{1}{2}(x^2+p_x^2)$ & \texttt{xQ*$p_x$Q+} & Yes\\\cline{2-4}
		& $H_2=\frac{1}{2}(4y^2+p_y^2)$ & \texttt{yQOO$p_y$Q+} & Yes \\\cline{2-4}
		& $H_3=x\sqrt{H_1H_2-l^2}-l p_x\ (l=xp_y-2yp_x)$ & \texttt{$H_1H_2$*lQ-Rx*l$p_x$*-} & No \\\hline
		\multirow{4}{*}{Three Body Problem}
		& $H_1=\sum_{i=1}^3 \frac{1}{2}(p_{i,x}^2+p_{i,y}^2)-(\frac{1}{r_{12}}+\frac{1}{r_{13}}+\frac{1}{r_{23}})$ & \texttt{$\sum_i p_{i,x}$Q$p_{i,y}$Q+$r_{i(i+1)}$IO-} &  Yes\\\cline{2-4}
		& $H_2=\sum_{i=1}^3 x_ip_{i,y}-y_ip_{i,x}$ & \texttt{$\sum_ix_ip_{i,y}$*$y_ip_{i,x}$*-} & Yes \\\cline{2-4}
		&$H_3=\sum_{i=1}^3 p_{i,x}$ & $\sum_ip_{i,x}$ & Yes \\\cline{2-4}
		&$H_4=\sum_{i=1}^3 p_{i,y}$ & $\sum_ip_{i,y}$ & Yes \\\hline
		\multirow{4}{*}{KdV} & $H_1=\int\phi\ dx $ & \texttt{$\phi$} & Yes \\\cline{2-4}
		& $H_2=\int\phi^2\ dx$ & \texttt{$\phi$Q} & Yes \\\cline{2-4}
		& $H_3=\int(2\phi^3-\phi_x^2)\ dx$ & \texttt{$\phi$Q$\phi$*O$\phi_x$Q-} & Yes
		\\\cline{2-4}
		&$H_4=\int(5\phi^4-10\phi\phi_x^2+\phi_{xx}^2)\ dx$ & \texttt{$\phi$QQ5*$\phi_x$Q$\phi$*10*-$\phi_{xx}$Q+} & No
		\\\hline
		\multirow{3}{*}{Nonlinear Schr\"odinger}
		& $H_1=\int|\psi|^2\ dx$ & \texttt{$\psi$Q} &  Yes \\\cline{2-4}
		& $H_2=\int(|\psi_x|^2+|\psi|^4)\ dx$ & \texttt{$\psi_x$Q$\psi$QQ+} &  Yes \\\cline{2-4}
		& $H_3=\int(|\psi_{xx}|^2+2|\psi_x|^2|\psi|^2-2|\psi|^6)\ dx$ & \texttt{$\psi_{xx}$Q$\psi$Q$\psi_x$QO*+$\psi$QQ$\psi$Q*O-} &  No
		\\\hline
	\end{tabular}
	\caption{16 of the 20 conservation laws were discovered not only numerically, but also symbolically using our fast-rejection brute force search limited to 9 distinct symbols.}
	\label{tab:symbolic}
	\vskip -0.2cm
\end{table*}

{\bf Summary of numerical experiments} We test AI Poincar\'{e} 2.0 on several systems: the Kepler problem, the damped harmonic oscillator, the isotropic/anisotropic harmonic oscillators , the gravitational three-body problem, the KdV wave equation and the nonlinear Schr{\"o}dinger equation. 
The neural network has 2 hidden layers, each containing 256 neurons with SiLU activation, and is trained with the Adam optimizer~\cite{kingma2014adam} for 100 epochs. When training multiple networks simultaneously, we choose the regularization coefficient $\lambda=0.02$. Our method succeeds in discovering all conservation laws numerically (FIG.~\ref{fig:examples}) and most symbolically (Table ~\ref{tab:symbolic}). Below we go through these examples one by one.

\subsection{2D Kepler Problem}
The 2D Kepler Problem is described by two coordinates $(x,y)$ and two velocity components $(v_x,v_y)$,
\begin{equation}\label{eq:kepler}
    \z=
    \begin{pmatrix}
    x \\ v_x \\ y \\ v_y
    \end{pmatrix}
    , \f(\z)=
    \begin{pmatrix}
    v_x \\ -GMx/(x^2+y^2)^{3/2} \\ v_y \\ -GMy/(x^2+y^2)^{3/2}
    \end{pmatrix}
\end{equation}
where $G$ is the gravitational constant, $M$ and $m$ are the mass of the sun and the planet, respectively.
The system has three conserved quantities: (1) energy $H_1=-GMm/\sqrt{x^2+y^2}+\frac{m}{2}(v_x^2+v_y^2)$; (2) angular momentum $H_2=m(xv_y-yv_x)$; (3) The direction of the Runge-lenz vector $H_3={\rm arctan}(\frac{v_x H_2+GM\hat{r}_y}{-v_yH_2+GM\hat{r}_x})$ where $\hat{r}\equiv (\hat{r}_x,\hat{r}_y)=(\frac{x}{\sqrt{x^2+y^2}},\frac{y}{\sqrt{x^2+y^2}})$. Without loss of generality, $GM=1$. As shown in FIG~\ref{fig:examples} first column, out method correctly identifies all of three conservation laws.

The reverse Polish notation for $\sqrt{x^2+y^2}$ is {\tt xQyQ+R} (6 symbols) which is quite expensive. To facilitate symbolic learning, one may wish to add in the radius variable $r=\sqrt{x^2+y^2}$ to exploit the symmetry of the problem. To do so, we augment the original system with the extra variable $r$ into an augmented system:

\begin{equation}\label{eq:kepler_2}
    \z'=
    \begin{pmatrix}
    x \\ v_x \\ y \\ v_y \\ r
    \end{pmatrix}
    , \f'(\z')=
    \begin{pmatrix}
    v_x \\ -GMx/(x^2+y^2)^{3/2} \\ v_y \\ -GMy/(x^2+y^2)^{3/2} \\
    (xv_x+yv_y)/r
    \end{pmatrix}
\end{equation}
Our method manages to rediscover the symbolic formulas for energy and angular momentum, but the one for the Runge-Lenz vector is too long to be discovered, as shown in Table \ref{tab:symbolic}.

\subsection{1D Damped Harmonic Oscillator}

\begin{figure*}
    \centering
    \includegraphics[width=1.0\linewidth, trim=0cm 0.0cm 0cm 0cm]{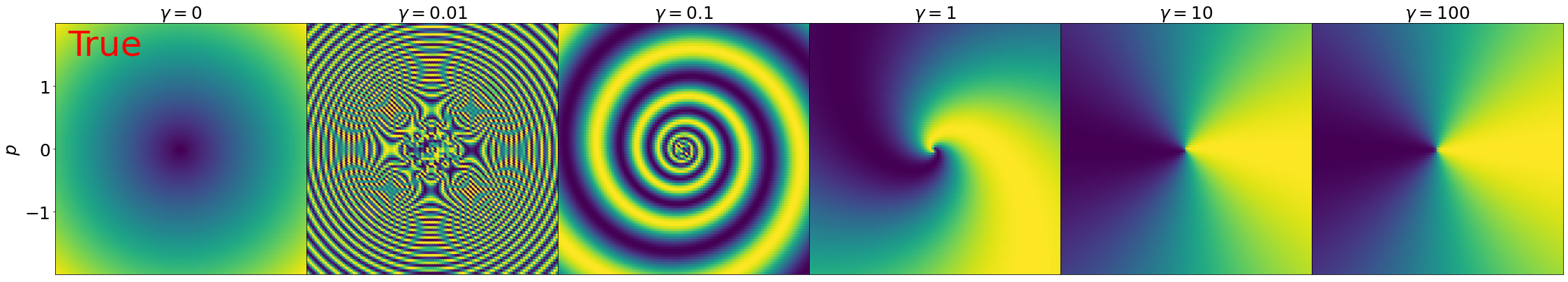}
    \includegraphics[width=1.0\linewidth, trim=0cm 0cm 0cm 0.5cm]{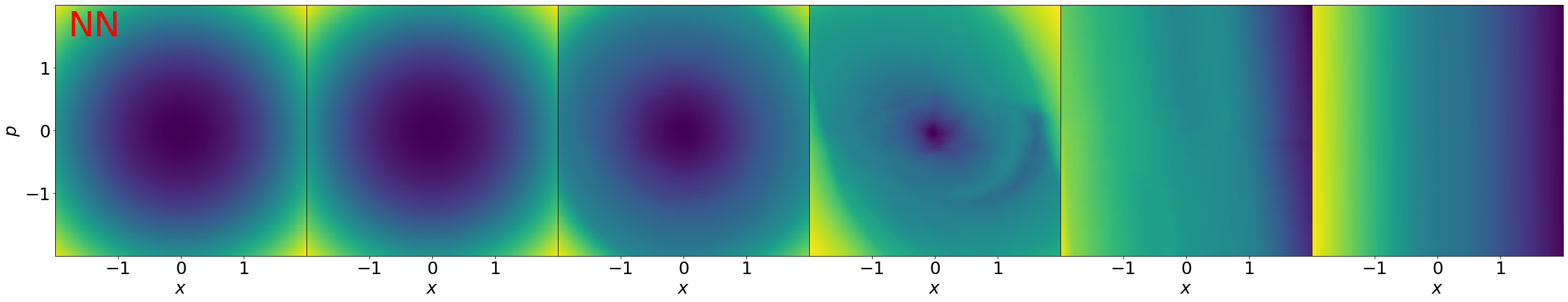}
    \includegraphics[width=0.95\linewidth, trim=0.5cm 0cm 2cm 0cm]{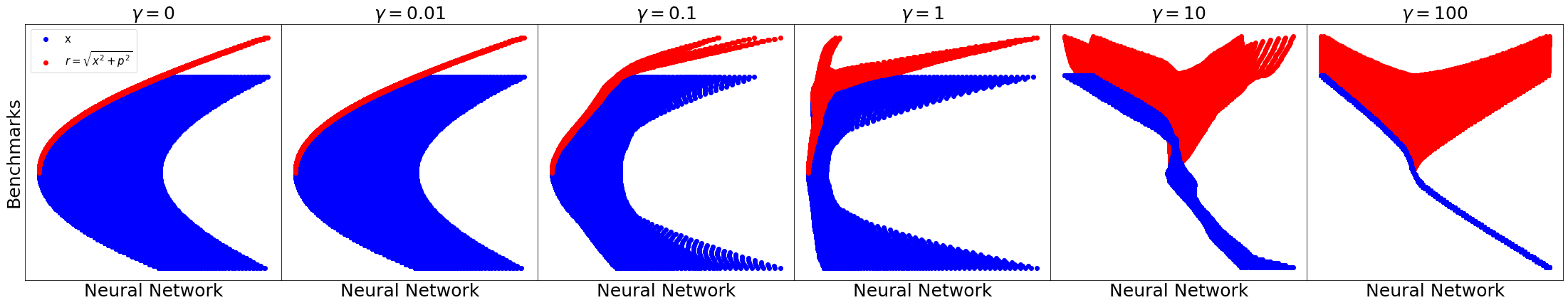}
    
    \caption{1D damped harmonic oscillator: Each column corresponds to a damping coefficient $\gamma$. Top: The conserved quantity of the 1D damped harmonic oscillator with different $\gamma$.
    Neural networks cannot perfectly learn the singular behavior near the origin, and also struggle when the stripes get too narrow. Middle: visualizations of neural network predictions of the conserved quantity. Bottom: Comparison of neural network predictions with $x$ and $r=\sqrt{x^2+p^2}$. For $\gamma=0$ and $\gamma=100$, the neural network learns $r$ and $x$ as conservation laws, respectively.}
    \label{fig:1d_ho_nn}
\end{figure*}

1D damped harmonic oscillator is described by the equation
\begin{equation}\label{eq:1d_damp}
    \frac{d}{dt}\begin{pmatrix}
    x \\ p
    \end{pmatrix} = 
    \begin{pmatrix}
    p \\ 
    -x - \gamma p 
    \end{pmatrix},
\end{equation}
where $\gamma$ is the damping coefficient. In the sense of Frobenius integrability (defined in Section \ref{app:integrability}), the system has 1 conserved quantity. We first attempt to construct the quantity analytically. The family of solutions for Eq.~(\ref{eq:1d_damp}) is 
\begin{equation}
    \begin{pmatrix}
    x(t) \\
    p(t)
    \end{pmatrix} = 
    \begin{pmatrix}
    e^{-\gamma t}{\rm cos}(t+\varphi) \\ 
    e^{-\gamma t}{\rm sin}(t+\varphi)
    \end{pmatrix}, \quad \varphi\in[0,2\pi).
\end{equation}
Define the complex variable $z(t)\equiv x(t)+ip(t)=e^{(-\gamma+i)t+i\varphi}$ and its complex conjugate $\bar{z}=e^{(-\gamma-i)t-i\varphi}$. Then

\begin{equation}
    H\equiv z^{(-\gamma-i)}/\bar{z}^{(-\gamma+i)}=\left(\frac{z}{\bar{z}}\right)^{-\gamma} (z\bar{z})^{-i}=e^{-2i \gamma\varphi}
\end{equation}
is a conserved quantity. When $\gamma=0$, $H\sim (z\bar{z})=|z|^2=x^2+p^2$ which is the energy; when $\gamma\to\infty$, $H\sim (z/\bar{z})\sim {\rm arg}(z)={\rm arctan}(p/x)$ which is the polar angle. For visualization purposes, we define $H'\equiv \frac{i}{2\gamma}{\rm ln}H=\theta+\frac{{\rm ln}r}{\gamma}$, where $\theta={\rm arctan}\frac{p}{x}$ and $r=\sqrt{x^2+p^2}$. We visualize ${\rm cos}H'$ in FIG.~\ref{fig:1d_ho_nn} top for different $\gamma$. The function looks regular for $\gamma=0$ and  $\gamma\geq 10$, but looks ill-behaved for e.g., $\gamma=0.01$ and $0.1$.

\begin{figure}
    \centering
    \includegraphics[width=1.0\linewidth]{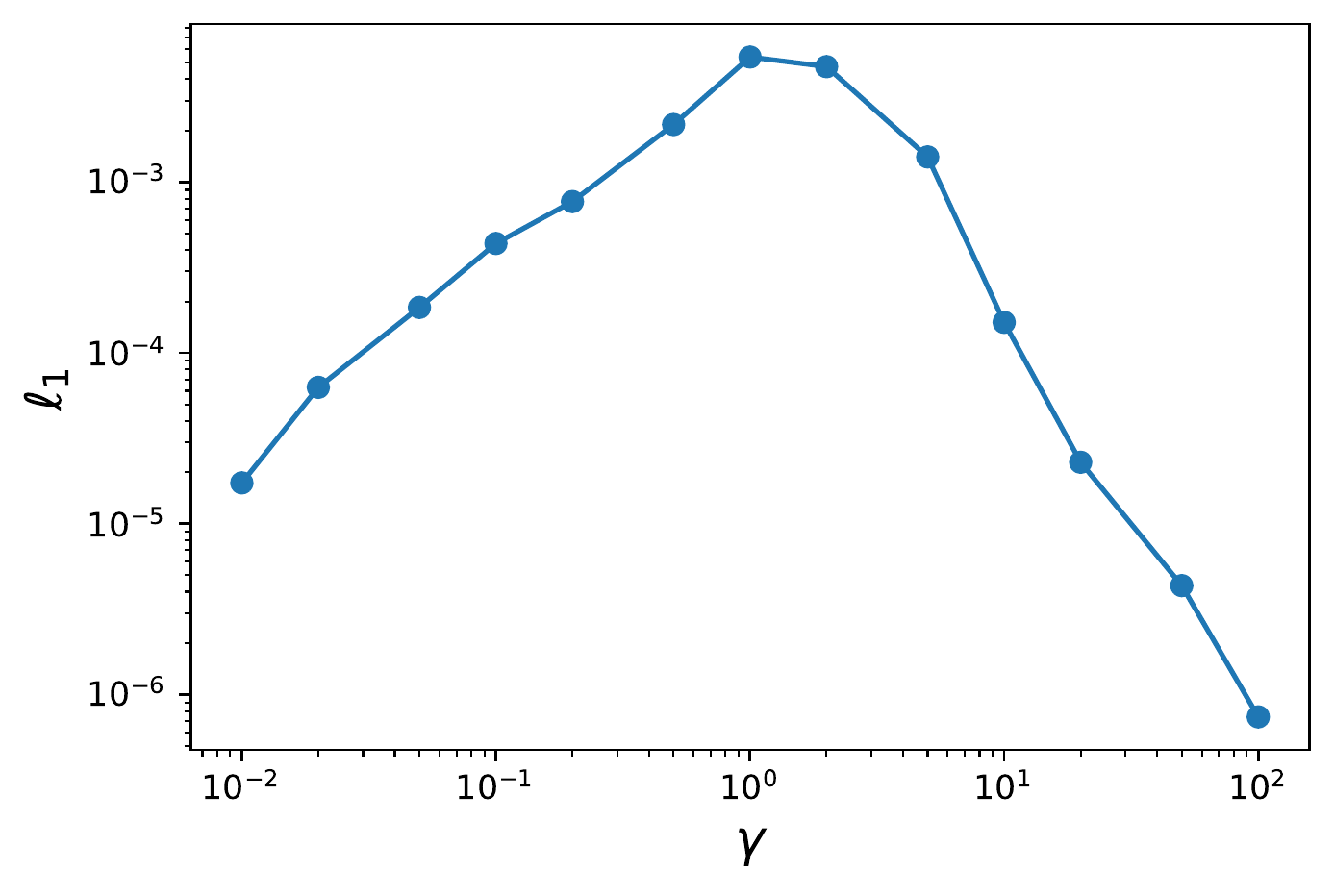}
    \caption{1D damped harmonic oscillator: conservation loss $\ell_1$ as a function of $\gamma$.}
    \label{fig:1d_ho_l1}
\end{figure}

{\bf Neural networks cannot learn ill-behaved conserved quantities well}. Neural networks have an implicit bias towards smooth functions, so they are unable to learn ill-behaved conserved quantities. To verify the argument, we run AI Poincar{\'e} 2.0 (only an $n=1$ model, hence no regularization) on the 1D damped harmonic oscillator with different damping coefficient $\gamma$, and plot $\ell_1$ as a function of $\gamma$ in FIG.~\ref{fig:1d_ho_l1}. We found that: (1) the conservation loss $\ell_1$ is almost vanishing at small $\gamma=0.01$ and large $\gamma=100$; (2) $\ell_1$ peaks around $\gamma=1$, which agrees with the visualization in FIG.~\ref{fig:1d_ho_nn} top row. We visualize functions learned by neural networks in FIG.~\ref{fig:1d_ho_nn} middle row, each column displaying results of a specific $\gamma$. To interpret what conserved quantity the neural network has learned, we compare the learned function $H(x,p)$ with two baseline functions $H_1(x,p)=r\equiv\sqrt{x^2+p^2}$ and $H_2(x,p)=x$ in FIG.~\ref{fig:1d_ho_nn} bottom row. If $H$ and $H_i (i=1, 2)$ are the same function up to an overall nonlinear transformation, i.e., $H=f(H_i)$, then 2D scatter points $(H(x,p), H_i(x,p))$ for all $(x,p)$ pairs should only occupy a 1D sub-manifold in 2D. When the scatter points do not have a submanifold structure, it implies that $H$ and $H_i$ are not the same function. When $\gamma=0.01$, the conserved quantity is equivalent to $r$ up to a nonlinear re-parameterization; When $\gamma=100$, the conserved quantity is equivalent to $x$ up to a nonlinear re-parameterization. 

While advanced techniques~\cite{sitzmann2020implicit} can bias neural networks towards highly oscillatory and/or ill-behaved functions, the smoothness of neural networks is a feature than bug for physicists who care about only well-behaved conserved quantities. We will expand on this idea in Section \ref{app:integrability}.

\subsection{2D Isotropic and Anisotropic Harmonic Oscillator}
The Harmonic Oscillator (2D) is described by two coordinates $(x,y)$ and two momenta $(p_x,p_y)$.
\begin{equation}
    \mat{z}=
    \begin{pmatrix}
    x \\ p_x \\ y \\ p_y
    \end{pmatrix}
    , \mat{f}(\z)=
    \begin{pmatrix}
    p_x/m \\ -\omega_x^2x \\ p_y/m \\ -\omega_y^2y
    \end{pmatrix},
\end{equation}
where $m$ is the mass, and $\omega_x$ and $\omega_y$ are angular frequencies. When $\omega_x\neq \omega_y$, the system is anisotropic and has two obvious conserved quantities: (1) $x$-energy $H_1=\frac{1}{2}\omega_x^2x^2+\frac{1}{2m}p_x^2$ and (2) $y$-energy $H_2=\frac{1}{2}\omega_y^2y^2+\frac{1}{2m}p_y^2$. The third conserved quantity is less studied by physicists but still exists if $\omega_x/\omega_y$ is a rational number~\cite{Arutyunov2019}. When $\omega_x=\omega_y$, the system is isotropic and has three conserved quantities. Besides $H_1$ and $H_2$, angular momentum $H_3=xp_y-yp_x$ is also conserved. For the isotropic case, we choose $m=\omega_x=\omega_y=1$; for the anisotropic case, we choose $m=\omega_x=1,\omega_y=2$. Samples are drawn from the uniform distribution $\z\sim U[-2,2]^4$. We include more physics discussion below for completeness.

{\bf Isotropic case} In the isotropic case $\omega_x=\omega_y=m=1$, there are four conservation laws~\cite{2dhoiso}:

\begin{equation}
    \begin{aligned}
    &2H_1=x^2+p_x^2,\quad 2H_2=y^2+p_y^2, \\ &L=yp_x-xp_y,\quad K = xy+p_xp_y.
    \end{aligned}
\end{equation}
but they are dependent because $L^2+K^2=4H_1H_2$. $H_1$, $H_2$ and $L$ are more common in physics, while $K$ is less common. However, there is no need to prefer $L$ over $K$. In fact, our symbolic module discovers the three conserved quantities $2H_1,2H_2,K$ and then ignores $L$ because of its dependence on the other three quantities, shown in Table \ref{tab:symbolic}. The ordering of $L$ and $K$ is in fact arbitrary. In terms of reverse polish notation, both $K=xy*p_xp_y*+$ and $L=yp_x*xp_y*-$ belong to the template $0020022$ where $0$ represents a variable and $2$ represents a binary operator. Because we try “$+$” before “$-$”, $K$ comes before $L$. If we instead try “$-$” before “$+$”, then $L$ comes before $K$. As a sanity check, our method discovered the correct number (3) of conservation laws, as shown in FIG.~\ref{fig:examples} second column.

{\bf Anisotropic case} Something amusing happened for the anisotropic oscillator example. The first author, despite passing his classical mechanics exam with full score, expected two IOMs rather than three because the angular momentum is not conserved for the anisotropic oscillator. However, AI Poincar{\'e} insisted there were three IOMs, as shown in FIG.~\ref{fig:examples} third column. The authors eventually realized that AI Poincar{\'e} was right: a third IOM is indeed present, although poorly known among physicists ~\cite{Arutyunov2019}.

Let us consider the specific case $m=\omega_x=1,\omega_y=2$. The equations of motion are:
\begin{equation}
\frac{d}{dt}
    \begin{pmatrix}
    x \\
    v_x \\
    y \\ 
    v_y \\
    \end{pmatrix}=
    \begin{pmatrix}
    v_x \\
    - x \\
    v_y \\
    - 4 y
    \end{pmatrix}.
\end{equation}
Solving the equation yields the trajectory
\begin{equation}
    \begin{pmatrix}
    x \\
    v_x \\
    y \\ 
    v_y \\
    \end{pmatrix}=
    \begin{pmatrix}
    A_x{\rm sin}( t+\varphi_x) \\
    A_x {\rm cos}( t+\varphi_x) \\
    A_y{\rm sin}(2 t+\varphi_y) \\
    2 A_y{\rm cos}(2 t+\varphi_y)
    \end{pmatrix}
\end{equation}
with arbitraty constants $A_x$, $A_y$, $\varphi_x$ and $\varphi_y$.

We define angular momentum
\begin{equation}
    L^{(1)}\equiv xp_y-yp_x
    = 2A_xA_y({\rm sin}(t+\varphi_1-\varphi_2)).
\end{equation}
Note $L^{(1)}$ is not conserved, nor is $K^{(1)}\equiv \sqrt{(2 A_xA_y)^2-L^{(1)2}}=2 A_xA_y{\rm cos}( t+\varphi_1-\varphi_2)$. However, it is interesting to note that the trajectory of $\z'\equiv(x,v_x,L^{(1)},K^{(1)})$ can be generated from an isotropic harmonic oscillator, because all components have the same angular frequency. Hence the `angular momentum' is conserved:
\begin{equation}
\begin{aligned}
&L^{(2)}\equiv xK^{(1)}-yL^{(1)}=\\
&x(xp_y-yp_x)-y\sqrt{(x^2+p_x^2)(y^2+p_y^2)-(xp_y-yp_x)^2}
\end{aligned}
\end{equation}
Although the numerical front realizes the existence of this conserved quantity, it remains difficult for the symbolic front to discover it due to its length, as shown in Table \ref{tab:symbolic}.

\begin{figure*}
    \centering
    \includegraphics[width=1\linewidth]{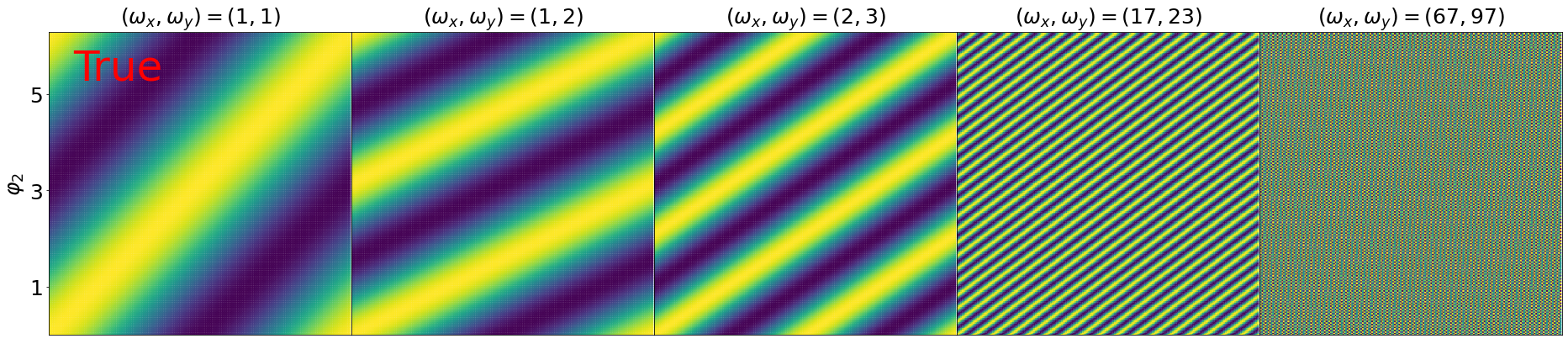}
    \includegraphics[width=1.0\linewidth, trim=0cm 0cm 0cm 0.5cm]{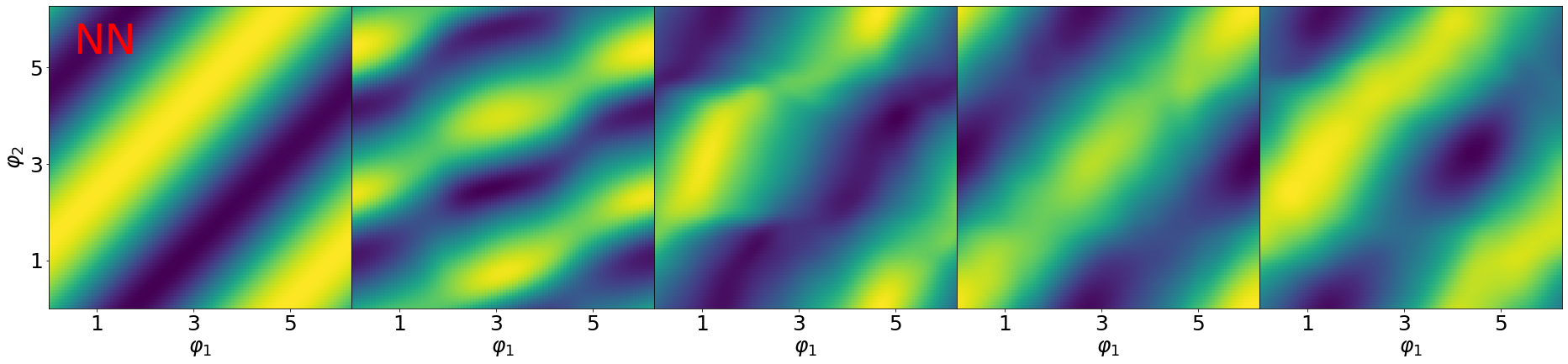}
    \caption{The third conserved quantity of the 2D harmonic oscillator with different frequency pairs $(\omega_x, \omega_y)$. Top: ground Truth; bottom: learned results by neural networks.
A neural network can only learn this conserved quantity if the frequency ratio $q\equiv\omega_x/\omega_y$  is a ratio of small integers; if $q$ is irrational, the conserved quantity is an everywhere discontinuous function that is completely useless to physicists. 
    }
    \label{fig:2d_ho_fractal}
\end{figure*}

For general $(\omega_x,\omega_y)$, there exists a third conserved quantity in the sense of Frobenius integrability, as we construct below (also in \cite{2dhoiso}). The family of solutions is 
\begin{equation}
    \begin{pmatrix}
    x \\
    p_x \\
    y \\
    p_y
    \end{pmatrix}=
    \begin{pmatrix}
    A_x{\rm cos}(\omega_x t+\varphi_x) \\
    -\omega_x A_x{\rm sin}(\omega_x t+\varphi_x) \\
    A_y{\rm cos}(\omega_y t+\varphi_y) \\
    -\omega_y A_y {\rm sin}(\omega_y t+\varphi_y)
    \end{pmatrix}
\end{equation}
We define $z_1\equiv \frac{1}{A_x}( x+i\frac{p_x}{\omega_x})=e^{i(\omega_x t+\varphi_x)}$, and $z_2\equiv \frac{1}{A_y}(y+i\frac{p_y}{\omega_y})=e^{i(\omega_y t+\varphi_y)}$. Hence 
\begin{equation}
    H_3 \equiv z_1^{\omega_y}/z_2^{\omega_x} = e^{i(\omega_y\varphi_x-\omega_x\varphi_y)}
\end{equation}
is a conserved quantity. In the isotropic case when $\omega_x=\omega_y=\omega$, $H_3$ simplifies to 
\begin{equation}
    H_3 = (\omega^2xy+p_xp_y+i\omega(xp_y-yp_x))/H_2
\end{equation}
whose imaginary part is the well-known angular momentum. Since the norm of $H_3$ is 1, the real and imaginary part are not independent. We plot $-i{\rm ln}H_3$ in FIG.~\ref{fig:2d_ho_fractal} top with different $(\omega_x,\omega_y)$. We set $A_x=A_y=1$. In the cases when $\omega_y/\omega_x$ is an integer or simple fractional number, $H_3$ is regular; however when $\omega_y/\omega_x$ is a complicated fractional number or even an irrational number, $H_3$ is ill-behaved, demonstrating fractal behavior.

We also run AI Poincar{\'e} 2.0 ($n=4$ models are trained) on the 2D harmonic oscillator example with different frequency ratios $\omega_y/\omega_x$. In FIG.~\ref{fig:2d_ho_fractal} bottom, we visualize the worst conserved quantity, i.e., the one with the highest conservation loss, out of 4 neural networks. To map the four-dimensional function to a 2D plot, we constrain $x = {\rm cos}\varphi_1, p_x={\rm sin}\varphi_1, y = {\rm cos}\varphi_2, p_y={\rm sin}\varphi_2$. When $(\omega_x,\omega_y)=(1,1)$ or $(1,2)$, the neural network prediction of the third conserved quantity aligns well with our expectation (visualized in FIG.~\ref{fig:2d_ho_fractal}). For more complicated $\omega_y/\omega_x$ ratios, the prediction looks similar to the $(\omega_x,\omega_y)=(1,1)$ case, but they have high conservation loss, as shown in TABLE \ref{tab:2d_ho_nn}.

\begin{table}[]
    \centering
    \resizebox{0.45\textwidth}{!}{\begin{tabular}{|c|c|c|c|c|c|}\hline
    $(\omega_x,\omega_y)$     & $(1,1)$ & $(1,2)$ & $(2,3)$ & $(17,23)$ & $(67,97)$ \\\hline
    Worst conservation loss    & $1.1\times 10^{-4}$ & $5.1\times 10^{-4}$ & $7.9\times 10^{-4}$ & $1.2\times 10^{-3}$ & $1.4\times 10^{-3}$ \\\hline
    Average conservation loss & $7.7\times 10^{-5}$ & $4.6\times 10^{-4}$ & $4.7\times 10^{-4}$ & $1.0\times 10^{-3}$ & $1.1\times 10^{-3}$ \\\hline
    \end{tabular}}
    \caption{2D harmonic oscillator: worst and average conservation loss for different ratios $\omega_y/\omega_x$.}
    \label{tab:2d_ho_nn}
\end{table}

\subsection{Three-body Problem}\label{sec:three_body}

The three-body problem has 12 degrees of freedom: 6 positions $(x_i,y_i) (i=1,2,3)$ and 6 velocities $(v_{x,i},v_{y,i}) (i=1,2,3)$.
Although there are 12-1=11 IOMs, only 4 are identified as conservation laws by physicists: (1) $x$-momentum: $H_1=\sum_{i=1}^3m_iv_{i,x}$; (2) $y$-momentum: $H_2=\sum_{i=1}^3 m_iv_{i,y}$; (3) angular momentum: $H_3=\sum_{i=1}^3 m_i(x_iv_{i,y}-y_iv_{i,x})$; (4) energy $H=\sum_{i=1}^3\frac{1}{2}m_i(v_{i,x}^2+v_{i,y}^2)+(\frac{Gm_1m_2}{((x_1-x_2)^2+(y_1-y_2)^2)^{1/2}}+\frac{Gm_1m_3}{((x_1-x_3)^2+(y_1-y_3)^2)^{1/2}}+\frac{Gm_2m_3}{((x_2-x_3)^2+(y_2-y_3)^2)^{1/2}})$. In numerical experiments, we set $G=m_1=m_2=m_3=1$. Similar to the Kepler problem, we can simplify symbolic search by adding three distance variables:
\begin{equation}
    \begin{aligned}
    r_{12} = \sqrt{(x_1-x_2)^2+(y_1-y_2)^2}, \\
    r_{13} = \sqrt{(x_1-x_3)^2+(y_1-y_3)^2}, \\
    r_{23} =
    \sqrt{(x_2-x_3)^2+(y_2-y_3)^2}. \\
    \end{aligned}
\end{equation}

According to Landau~\cite{landau1976mechanics}, conservation laws are those IOMs which respect spacetime symmetries and being additive. To incorporate these inductive biases, we assume that a conserved quantity decomposes into 1-body terms and 2-body terms. 
We assume nothing about the 1-body terms, but assume translational and rotational invariance for the 2-body terms. As a result, a candidate conservation law must have the form:
\begin{equation}
    H = \sum_{i=1}^3 g(x_i,y_i,v_{i,x},v_{i,y})+\sum_{i=1}^3\sum_{j=i+1}^3 h(r_{ij})
\end{equation}
where $r_{ij}\equiv \sqrt{(x_j-x_i)^2+(y_j-y_i)^2}$. By parameterizing $g$ and $h$ as two separate neural networks, the learned conservation laws automatically satisfy the above-mentioned desired physical properties.  Our algorithm now discovers precisely 4 independent conservation laws, as shown in FIG.~\ref{fig:examples} fourth column. 

It is useful to push the limit of our method to see it still works in more challenging scenarios. We investigate two cases below: (1) no inductive biases or (2) unequal masses.

\begin{figure}
    \centering
    \includegraphics[width=1.0\linewidth]{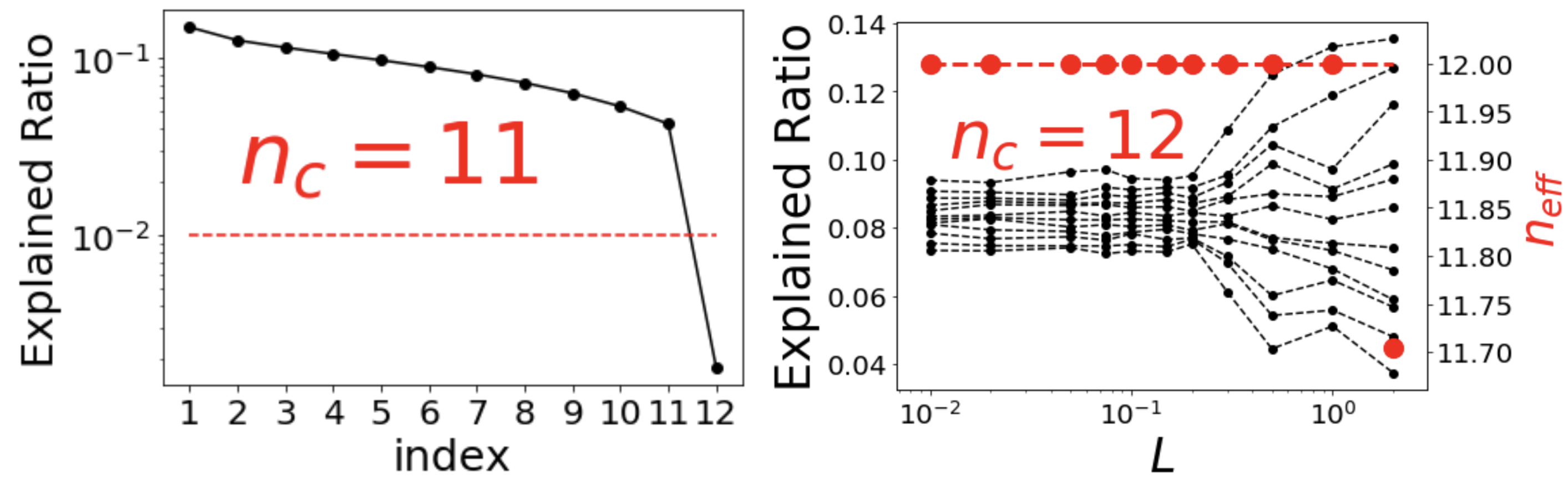}
    \caption{The 2D three body problem without inductive biases. The differential rank (left) correctly predicts 11 IOMs, while the rank (right) incorrectly predicts 12 IOMs. This implies that the differential rank is preferred over the rank in high dimensions, i.e., when $n_c$ is large.}
    \label{fig:threebody_noib}
\end{figure}

{\bf Challenging case 1: No inductive biases.} When no inductive bias is added to the neural network, the neural network degrades to parameterize integrals of motion. Since a first-order differential equation with $s$ degrees of freedom have $s-1$ integrals of motion, the 2D three-body problem has $12-1=11$ integrals of motion. The results are quite interesting: the differential rank method predicts correctly 11 IOMs (FIG.~\ref{fig:threebody_noib} left), while the rank method predicts incorrectly 12 IOMs (FIG.~\ref{fig:threebody_noib} right). This is possibly because Neural Empirical Bayes (the manifold learning module used to compute rank, as well as in AI Poincar\'{e} 1.0) degrades when dealing with high-dimensional manifolds. This highlights yet another benefit of differential rank, which is novely proposed in 2.0. Differential rank is not only more numerically efficient than rank, but also more stable in high dimensions.

\begin{figure}
    \centering
    \includegraphics[width=1.0\linewidth]{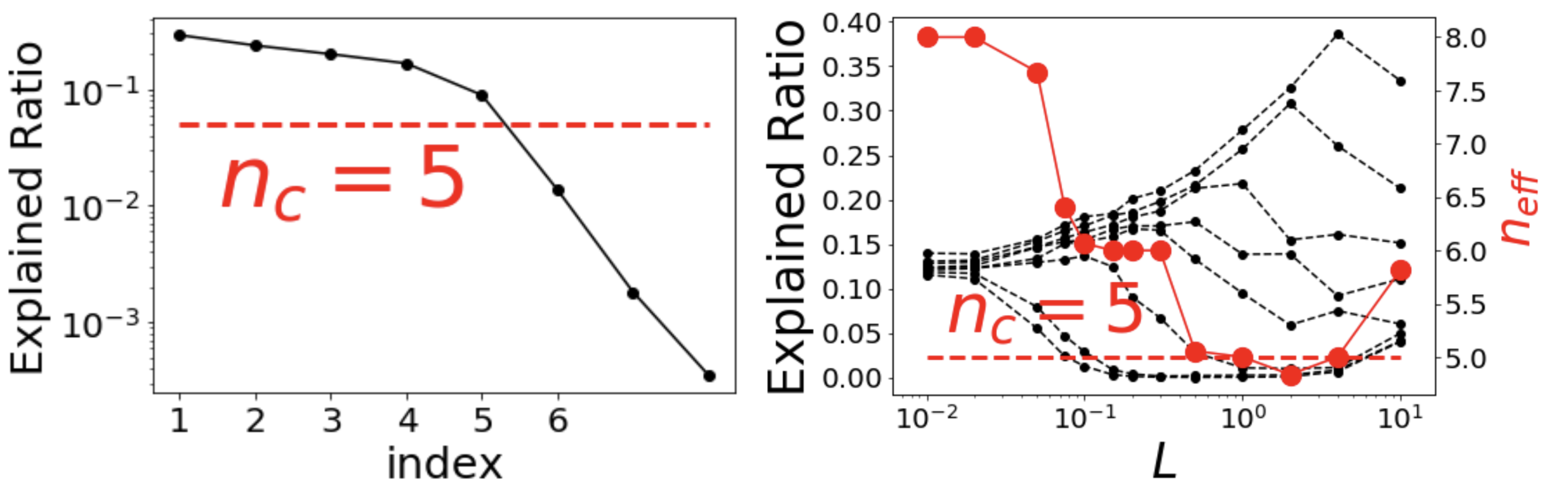}
    \caption{The 2D three body problem with uneuqal masses $m_1:m_2:m_3=400:20:1$. Both the differential rank (left) and the rank (right) correctly predict $n_c=5$ conservation laws. The result is different from $n_c=4$ for the equal masses case in FIG.~\ref{fig:examples}, implying that our method can also capture approximate conservation laws besides exact conservation laws.}
    \label{fig:threebody_unequal}
\end{figure}

{\bf Challenging case 2: Unequal masses} We tried a case in which $m_1:m_2:m_3 = 400:20:1$. Both the rank and the differential rank predict 5 conservation laws, shown in FIG.~\ref{fig:threebody_unequal} left and right. Interestingly, this is different from 4 conservation laws in the case of equal masses. We conjecture that this is because in the limit $m_1 \gg m_2 \gg m_3$: (1) the momentum of $m_1$ is almost conserved (2 conservation laws); (2) $m_2$ orbits around $m_1$ as in the Kepler problem (3 conservation laws); (3) any term involving $m_3$ can be ignored. So there are 2+3=5 conservation laws in total. The discrepancy between cases of equal or unequal masses is arguably a feature rather than a bug, implying that our method not only applies to \textit{exact} conservation laws, but also to \textit{approximate} ones.

\subsection{KdV Wave Equation}
Another set of interesting systems are \textit{partial differential equations} (PDE) in the form $u_t=f(u,u_x,u_{xx},\cdots)$. Since a field has infinite number of degrees of freedom (hence infinitely many IOMs), it is crucial to constrain the form of conservation laws to exclude trivial ones. In quantum mechanics, for example, any projector onto an eigenstate is an IOM, but these are less profound than probability conservation (known as unitarity) and energy conservation etc. Thus we focus on conservation laws with an integral form obeying translational invariance:
\begin{equation}
    H=\int h(u,|u_x|,|u_{xx}|,\cdots)\ dx
\end{equation}
In practice, we replace the integral by a sum over the points on a uniform grid.
Moreover, we take the absolute value of derivatives as inputs, \eg, $|u_x|$ and $|u_{xx}|$, to avoid trivial ``conserved quantities'' of the total derivative form $h={d\over dx} F(u,u_x,u_{xx},...)$, 
\eg, $u_x$, $u u_x$, or  $u_{xx}$, 
which are conserved simply due to zero boundary conditions.

The Korteweg–De Vries (KdV) equation is a mathematical model for shallow water surfaces. It is a nonlinear partial differential equation for a function $\phi$ with two real variables, $x$ (space) and $t$ (time):
\begin{equation}\label{eq:KdV}
    \phi_t + \phi_{xxx} - 6\phi\phi_x = 0.
\end{equation}
Zero boundary conditions are imposed at the ends of the interval $[a,b]$. The KdV equation is known to have infinitely many conserved quantities~\cite{KdV_cq}, which can be written explicitly as
\begin{equation}
    \int_{a}^{b} P_{2n-1}(\phi,\phi_x,\phi_{xx},\cdots)dx,
\end{equation}
which follows from locality and translational symmetry. The polynomials $P_n$ are defined recursively by 
\begin{equation}
    \begin{aligned}
    &P_1 = \phi, \\
    &P_n = -\frac{dP_{n-1}}{dx} + \sum_{i=1}^{n-2} P_iP_{n-1-i}.
    \end{aligned}
\end{equation}
The first few conservation laws are
\begin{equation}
    \begin{aligned}
    &\int \phi dx\quad\quad\quad\quad\quad{\rm \ (mass)}\\
    &\int \phi^2 dx \quad\quad\quad\quad\quad{\rm (momentum)} \\
    &\int (2\phi^3-\phi_x^2) dx \quad\ \ {\rm (energy)}
    \end{aligned}
\end{equation}
Despite infinitely many conservation laws, useful ones in physics are usually constrained to contain only $\phi$ and low-order derivatives $(\phi_x,\phi_{xx},\cdots)$. 
 

{\bf Converting to the canonical form $\dot{\z}=\f(\z)$}
Since our framework can only deal with systems with finite degrees of freedom, we need to discretize space. We discretize the interval $x\in[-10,10]$ uniformly into $N_p=40$ points, denoted $x_1,\cdots,x_{N_p}$ and only store derivatives up to fifth order on each grid point, using them to parametrize our $\phi(x)$. 
This transforms our PDE into an ordinary differential equation with $3N_p$ degrees of freedom ($\phi^{(i)}=\phi(x_i),\phi_x^{(i)}=\phi_x(x_i),\cdots$):
Eq.~(\ref{eq:KdV}) implies that
\begin{equation}
    \begin{aligned}
    \partial_t
    \begin{pmatrix}
        \phi \\
        \phi_x \\
        \phi_{xx} \\
        \vdots 
    \end{pmatrix} = 
    \begin{pmatrix}
    -\phi_{xxx} + 6\phi\phi_x \\
    -\phi_{xxxx} + 6(\phi_x^2+\phi\phi_{xx}) \\
    -\phi_{xxxxx} + 6(3\phi_x\phi_{xx}+\phi\phi_{xxx})\\
    \vdots
    \end{pmatrix}
    \end{aligned}
\end{equation}
so our discretized PDE problem becomes 
\begin{equation}
    \resizebox{0.5\textwidth}{!}{$\z\equiv
    \begin{pmatrix}
    \phi^{(1)} \\
    \phi_x^{(1)} \\
    \phi_{xx}^{(1)} \\
    \vdots \\
    \phi^{(N_p)} \\
    \phi_x^{(N_p)} \\
    \phi_{xx}^{(N_p)} \\
    \end{pmatrix},
    \f(\z)\equiv \partial_t \z =
    \begin{pmatrix}
    -\phi_{xxx}^{(1)} + 6\phi^{(1)}\phi_x^{(1)} \\
    -\phi_{xxxx}^{(1)} + 6(\phi_x^{(1)2}+\phi^{(1)}\phi_{xx}^{(1)}) \\
    -\phi_{xxxxx}^{(1)} + 6(3\phi_x^{(1)}\phi_{xx}^{(1)}+\phi^{(1)}\phi_{xxx}^{(1)})\\
    \vdots \\
    -\phi_{xxx}^{(N_p)} + 6\phi^{(N_p)}\phi_x^{(N_p)} \\
    -\phi_{xxxx}^{(N_p)} + 6(\phi_x^{(N_p)2}+\phi^{(N_p)}\phi_{xx}^{(N_p)}) \\
    -\phi_{xxxxx}^{(N_p)} + 6(3\phi_x^{(N_p)}\phi_{xx}^{(N_p)}+\phi^{(N_p)}\phi_{xxx}^{(N_p)})\\
    \end{pmatrix}$}
\end{equation}

{\bf Sample generation} We represent $\phi$ as a Gaussian mixture, so all derivatives can be computed analytically. In particular,
\begin{equation}
    \phi(x) = \sum_{i=1}^{N_g} A_i(\frac{1}{\sqrt{2\pi}\sigma_i}{\rm exp}(-(x-\mu_i)^2)/2\sigma_i^2), -10\leq x\leq 10
\end{equation}
where coefficients are set or drawn randomly accordingly to $A_i\sim U[-5,5]$, $\mu_i\sim U[-3,3], \sigma_i=1.5$. These distributions are chosen such that (1) $\phi(x)$ is (almost) zero at two boundary points $x=-10,10$; and (2) every single term in $\f(\z)$ have similar magnitudes. We choose $N_g=5$ and generate $P=10^4$ profiles of $\phi$.

{\bf Constraining conservation laws} The conservation laws of partial differential equations usually have the integral form, i.e., $H =\int h(x') dx$ where $x'=(\phi,\phi_x,\phi_{xx},\cdots)$. When space is discretized, we constrain the conservation law to the form $H=\sum_{i=1}^{N_p}h(x')$. On the numerical front, we parameterize $h(x')$ (as opposed to $H$) by a neural network; On the symbolic front, we search the symbolic formula of $h(x')$ (as opposed to $H$). The summation operation is hard coded for both fronts.

{\bf Avoiding trivial conservation laws} Due to zero boundary conditions, if $h(x')$ is an $x$-derivative of another function $g(x')$, then it is obvious that $\int_a^b h(x')dx = g(x')|_b - g(x')|_a=0$ which is a trivial conserved quantity. For example, $h(x')=\phi_x,\phi\phi_x,\phi_{xx},\phi_x^2+\phi\phi_{xx}$ are all trivial. We observe that each of them has at least one term that is an odd function of a derivative. Consequently a simple solution is to use absolute values $(|\phi_x|,|\phi_{xx}|,\cdots)$ instead of $(\phi_x,\phi_{xx},\cdots)$ so that these trivial conservation laws are avoided in the first place.

On the numerical front, our algorithm successfully discovers 2, 3, 4 conserved quantities which are dependent on $\phi$, $(\phi,\phi_x)$ and $(\phi,\phi_x,\phi_{xx})$ respectively, as shown in FIG.~\ref{fig:examples} second to last column. On the symbolic front, we constrain the input variables to be $(\phi,\phi_x,\phi_{xx})$, and three out of four conservation laws (mass, momentum and energy) can be discovered, as shown in Table \ref{tab:symbolic}. Our method fails for the fourth conservation law because it is too long.

\subsection{Nonlinear Schr\"odinger Equation}

The 1D nonlinear Schr{\"o}dinger equation (NLS) is a nonlinear generalization of the Schr{\"o}dinger equation. Its principal applications are to the propagation of light in nonlinear optical fibres and planar waveguides and to Bose-Einstein condensates. The classical field equation (in dimensionless form) is
\begin{equation}
    i\psi_t = -\frac{1}{2}\psi_{xx} + \kappa |\psi|^2\psi.
\end{equation}
Zero boundary conditions are imposed at infinity~\cite{Barrett2013TitleT}. Like the KdV equation, the NLS has infinitely many conserved quantities of the integral form
\begin{equation}
    H(x) = \int_{-\infty}^{\infty} h(\psi,\psi_x,\psi_{xx},\cdots) dx.
\end{equation}
Useful conservation laws in physics usually contain only low-order derivatives, e.g.,

\begin{equation}
    \begin{aligned}
    &{\rm unitarity:} \int |\psi|^2 dx \\
    &{\rm energy}: \int\frac{1}{2} \left(|\psi_x|^2+\kappa|\psi|^4\right) dx
    \end{aligned}
\end{equation}

{\bf Converting to the canonical form $\dot{\z}=\f(\z)$} Similar to the KdV equation, we treat $(\psi,\psi_x,\psi_{xx},\cdots)$ as different variables. We denote $\psi_{r} \equiv {\rm Re}(\psi), \psi_{i} \equiv {\rm Im}(\psi), {\rm Re}(\psi_x)=\psi_{x,r}, {\rm Im}(\psi_x)=\psi_{x,i}$, etc.
\begin{equation}
    \resizebox{0.5\textwidth}{!}{$\partial_t
    \begin{pmatrix}
    \psi \\
    \psi_x \\
    \psi_{xx} \\
    \vdots
    \end{pmatrix}=
    \begin{pmatrix}
    \frac{1}{2}i\psi_{xx}-i\kappa|\psi|^2\psi \\
    \frac{1}{2}i\psi_{xxx} - i\kappa(|\psi|^2\psi_{x}+(\psi_r\psi_{x,r}+\psi_i\psi_{x,i})\psi) \\ 
    \frac{1}{2}i\psi_{xxxx} - i\kappa(|\psi|^2\psi_{xx}+2(\psi_r\psi_{x,r}+\psi_i\psi_{x,i})\psi_x+(\psi_{x,r}^2+\psi_r\psi_{xx,r}+\psi_{x,i}^2+\psi_i\psi_{xx,i})\psi) \\
    \vdots
    \end{pmatrix}$}
\end{equation}
Since $\psi$ is a complex number, we should treat real and imaginary parts separately.
\begin{equation}
    \resizebox{0.5\textwidth}{!}{$\partial_t
    \begin{pmatrix}
    \psi_r \\
    \psi_i \\
    \psi_{x,r} \\
    \psi_{x,i} \\
    \psi_{xx,r} \\
    \psi_{xx,i} \\
    \vdots
    \end{pmatrix}=
    \begin{pmatrix}
    -\frac{1}{2}\psi_{xx,i}+\kappa|\psi|^2\psi_i \\
    \frac{1}{2}\psi_{xx,r}-\kappa|\psi|^2\psi_r \\
    -\frac{1}{2}\psi_{xxx,i} + \kappa(|\psi|^2\psi_{x_i}+(\psi_r\psi_{x,r}+\psi_i\psi_{x,i})\psi_i) \\ 
    \frac{1}{2}\psi_{xxx,r} - \kappa(|\psi|^2\psi_{x,r}+(\psi_r\psi_{x,r}+\psi_i\psi_{x,i})\psi_r) \\ 
    -\frac{1}{2}\psi_{xxxx,i} + \kappa(|\psi|^2\psi_{xx,i}+2(\psi_r\psi_{x,r}+\psi_i\psi_{x,i})\psi_{x,i}+(\psi_{x,r}^2+\psi_{r,i}\psi_{xx,r}+\psi_{x,i}^2+\psi_i\psi_{xx,i})\psi_i) \\
    \frac{1}{2}\psi_{xxxx,r} - \kappa(|\psi|^2\psi_{xx,r}+2(\psi_r\psi_{x,r}+\psi_i\psi_{x,i})\psi_{x,r}+(\psi_{x,r}^2+\psi_r\psi_{xx,r}+\psi_{x,i}^2+\psi_i\psi_{xx,i})\psi_r) \\
    \vdots
    \end{pmatrix}$}
\end{equation}
Just as in the KdV example, to avoid trivial solutions, we consider only the equations for magnitude $(|\psi|,|\psi_x|,|\psi_{xx}|,\cdots)$.
\begin{equation}
    \partial_t
    \underbrace{\begin{pmatrix}
    |\psi| \\
    |\psi_x| \\
    |\psi_{xx}| \\
    \vdots 
    \end{pmatrix}}_{\z} = 
    \underbrace{\begin{pmatrix}
    (\psi_r\partial_t\psi_{r}+\psi_i\partial_t\psi_{i})/|\psi| \\
    (\psi_{x,r}\partial_t\psi_{x,r}+\psi_{x,i}\partial_t\psi_{x,i})/|\psi_x| \\
    (\psi_{xx,r}\partial_t\psi_{xx,r}+\psi_{xx,i}\partial_t\psi_{xx,i})/|\psi_{xx}| \\
    \vdots
    \end{pmatrix}}_{\f}
\end{equation}

{\bf Sample generation} is similar to the KdV equations, with the only difference that real and imaginary parts are both treated as (independent) Gaussian mixtures.

We feed the neural network with (1) $\psi$ only; (2) $\psi$ and $|\psi_x|$; (3) $\psi$, $|\psi_x|$ and $|\psi_{xx}|$, and our method predicts 1, 2 and 3 conservation laws respectively (shown in FIG.~\ref{fig:examples} last column), which basically agree with the ground truth, although our method is unable to discover the momentum which involves $\psi_x$ because the input $|\psi_x|$ lacks the phase information. We would like to investigate how to include the phase information with the help of complex neural networks in future works.

\section{Discussion}

\subsection{Definitions of integrability and relations to AI Poincar{\'e} 1.0/2.0}
\label{app:integrability}

Conservation laws are closely related to the notion of integrability~\footnote{Informally speaking, an integrable system is a dynamical system with sufficiently many conserved quantities.}, which in turn has various definitions from different perspectives~\cite{enwiki:1058752403,vickers_2001}. Here we list five definitions of integrability and corresponding definitions of conserved quantities.

{\bf (1) General integrability [global geometry/topology]}. In the context of differential dynamical systems, the notion of integrability refers to the existence of an invariant regular foliation of phase space~\cite{enwiki:1058752403}. Consequently, a conserved quantity should be a well-behaved function globally, not demonstrating any fractal or other pathological behavior.

{\bf (2) Frobenius integrability [local geometry/topology]}. A dynamical system is said to be Frobenius integrable if, locally, the phase space has a foliation of invariant manifolds~\cite{enwiki:1058752403}. One major corollary of the Frobenius theorem is that a first-order dynamical system with $s$ degrees of freedom always has $s-1$ (local) integrals of motion. Consequently, a conserved quantity in the sense of Frobenius integrability does not require the foliation to be regular in the global sense. The visual differences between local and global conserved quantities are shown in FIG.~\ref{fig:1d_ho_nn}, and \ref{fig:2d_ho_fractal}.

{\bf (3) Liouville integrability [algebra]}. In the special setting of Hamiltonian systems, we have Liouville integrability, which focuses on algebraic properties of a Hamiltonian system~\cite{Arutyunov2019}. Liouville integrability states that there exists a maximal set of Poisson commuting invariants, corresponding to conserved quantities. A system in the $2n$-dimensional phase space is Liouville integrable if it
has $n$ independent conserved quantities which commute with each other, i.e., $\{H_i,H_j\}=0$. According to the Liouville-Arnold theorem~\cite{Arutyunov2019}, such systems can be solved exactly by quadrature, which is a special case of solvable integrability (the fifth criterion below).

{\bf (4) Landau integrability [concept simplicity]} Landau stated in his textbook~\cite{landau1976mechanics} that physicists prefer symmetric and additive IOMs and promote them as fundamental ``conservation laws".

{\bf (5) Solvable integrability [symbolic simplicity]}. Solvable integrability requires the determination of solutions in an explicit functional form~\cite{vickers_2001}. This property is intrinsic, but can be very useful to simplify and theoretically understand problems.

{\bf (6) Experimental integrability [robustness]}. In physics, we consider a conserved quantity useful if a measurement of it at some time $t$ can constrain the state at some later time $t'>t$. In experimental physics, a measurement of a physical quantity always contains some finite error. Hence a useful conserved quantity must not be infinitely sensitive to measurement error. In contrast, FIG.~\ref{fig:2d_ho_fractal} (top row) shows that, although a conserved quantity exists for all possible frequency pairs $(\omega_x,\omega_y)$, their robustness to noise differ widely. Once the noise scale significantly exceeds the width of stripe pattern, an accurate measurement of the conserved quantity is impossible, and a measurement of the ``conserved quantity" provides essentially zero useful information for predicting the future state. When the frequency ratio is an irrational number, 
the ``conserved quantity" becomes discontinuous and pathological throughout phase space and completely useless for making physics predictions.
This experimental integrability criterion is thus compatible with general integrability, not Frobenius integrability.

In summary, the various notions of integrability are used to study dynamical systems, but have different motivations and scopes. General integrability and Frobenius integrability characterize global and local geometry; Liouville integrability takes an algebraic perspective and applies only to Hamiltonian systems; Landau and solvable integrability instead focus on simplicity based on concepts and symbolic equations, respectively. To the best of our knowledge, there is no agreement on whether one particular definition outperforms others in all senses. We believe they are complementary to each other, rather than being contradictory or redundant. In AI Poincar{\'e} 1.0~\cite{poincare} and 2.0 (the current paper), we mostly did not mentioned explicitly which sense of integrability/conserved quantities we referred to. Fortunately, AI Poincar{\'e} 2.0 can flexibly adapt to all definitions, as summarized in Table \ref{tab:integrability}.

\begin{table}[]
    \centering
    \begin{tabular}{|c|c|c|c|c|c|}\hline
    &General & Frobenius & Liouville & Landau & solvable  \\\hline
    Poincar{\'e} 1.0 & Yes & No & No & No & Yes \\\hline
    Poincar{\'e} 2.0 & Yes & Yes & Yes~\footnote{This case is not included in paper, but is doable when we combine the techniques of searching for hidden symmetries in ~\cite{liu2021machine}.} & Yes & Yes \\\hline
    \end{tabular}
    \caption{Five integrability definitions and whether AI Poincar{\'e} 1.0/2.0 can deal with them.}
    \label{tab:integrability}
\end{table}

AI Poincar{\'e} 1.0 defines a trajectory manifold, which is orthogonal to the invariant manifold. The trajectory manifold is globally defined, and its dimensionality is a topological invariant. As a consequence, in AI Poincar{\'e} 1.0, conserved quantities satisfy general integrability. The symbolic part of AI Poincar{\'e} 1.0 looks for formulas with simple symbolic forms, in the spirit of solvable integrability.

AI Poincar{\'e} 2.0 addresses the problem of finding a maximal set of independent conserved quantities, in analogy to the goal of the Frobenius theorem~\cite{enwiki:ft} which searches for a maximal set of solutions of a regular system of first-order linear homogeneous partial differential equations. The loss formulation in Eq.~(\ref{eq:loss}) can be viewed as a variational formulation of the system of PDEs to be satisfied for conserved quantities. Consequently, AI Poincar{\'e} 2.0 (neural network front) is aligned with Frobenius integrability if there is only one training sample $\z$. In the presence of many training samples over the phase space, our algorithm becomes aligned with the notion of the general integrability, because the conserved quantity is parameterized as a neural network which has an implicit bias towards smooth and regular functions globally. Although we did not explicitly deal with Liouville integrability in this paper, the algebraic nature of Liouville integrability makes it simply a ``hidden symmetry problem" that is defined and solved by ~\cite{liu2021machine}, and the techniques in the current paper can further improve the process by determining functional dependence among invariants learned by neural networks.  The symmetry and additivity in Landau integrability is known in the machine learning literature as physical {\it inductive biases}, which can be elegantly handled by adding constraints to the architectures or loss functions~\cite{karniadakis2021physics,liu2021physics}. Finally, the symbolic front of AI Poincar{\'e} 2.0 addresses the problem of finding conserved quantities with simple symbolic formulas.


\subsection{Phase transitions and how to choose $\lambda$ }\label{app:lamb_pt}
Eq.~(\ref{eq:loss}) has a hyperparameter, the regularization coefficient $\lambda$. If $\lambda$ is too small, then multiple networks may learn dependent conserved quantities. If $\lambda$ is too large, then the regularization loss dominates the conservation loss, making the conservation laws inaccurate. As we argue below, the proper choice of $\lambda$ has a lower bound which is determined by the approximation error tolerance $\epsilon$, and an upper bound $O(1)$.

We first use two analytic toy examples to provide insight. In both cases, the number of neural networks $n$ is equal to the dimension $s$ of the problem, just to demonstrate all possible phase transitions. In practice, it is sufficient to choose $n=s-1$. The geometric intuition for minimizing the loss function Eq.~(\ref{eq:loss}) is that $\ell_1$ encourages $\nabla H_i$ to be orthogonal to $\f$ while the regularization loss $\ell_2$ encourages $\nabla H_i$ and $\nabla H_j$ $(j\neq i)$ to be orthogonal. 

\begin{figure}
    \centering
    \includegraphics[width=0.9\linewidth]{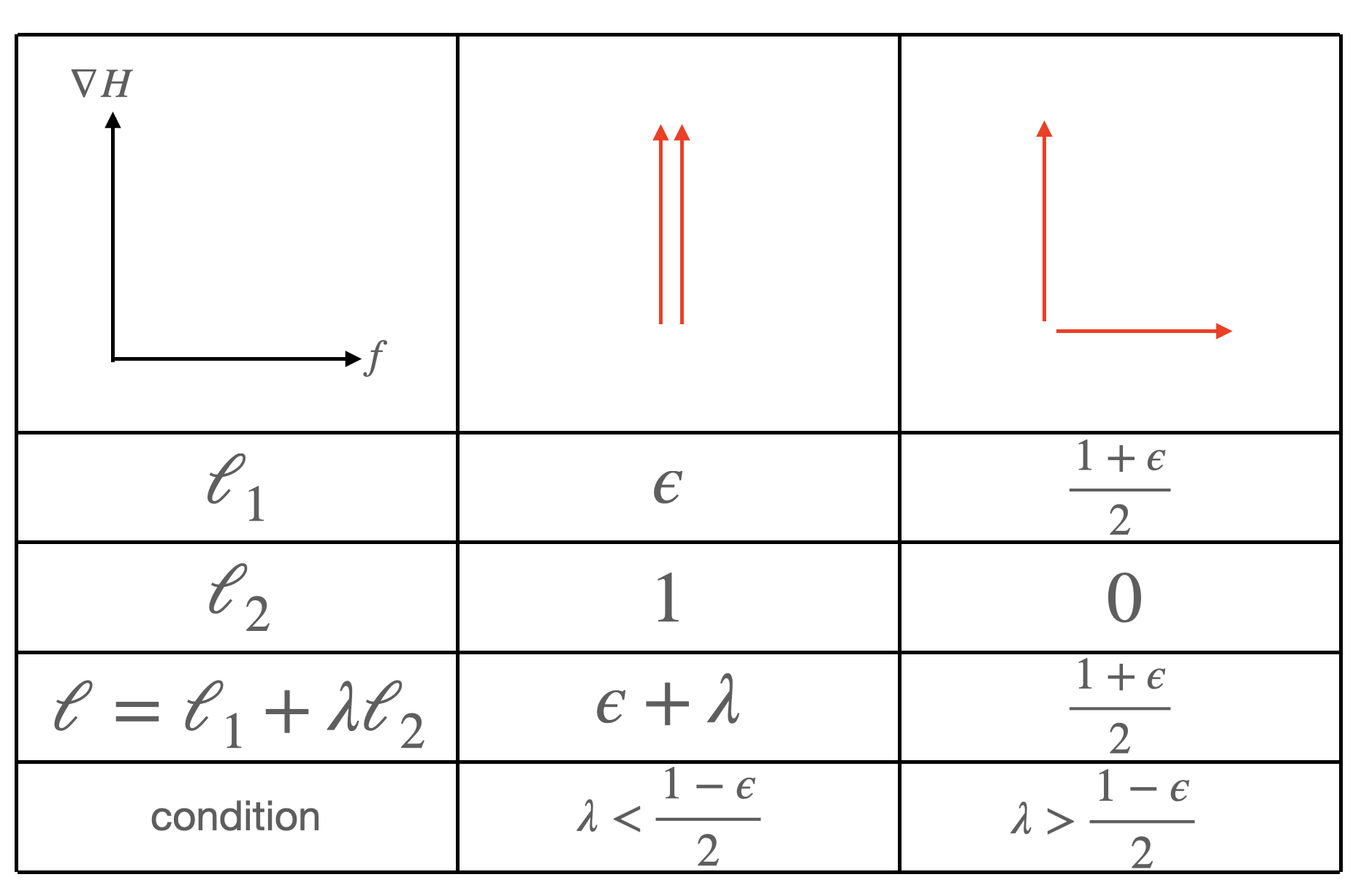}
    \caption{2D Toy example: With different $\lambda$, the global minima may have different geometric configurations. Assume the single conserved quantity can be approximated by a neural network with error $\epsilon$.}
    \label{fig:pt_toy1}
\end{figure}

{\bf Toy example 1:} The first toy example is inspired by the 1D damped harmonic oscillator with its 2D phase space. There is only one conserved quantity in the sense of Frobenius integrability, and the approximation error of a neural network is $\epsilon$. We train 2 networks to learn the conserved quantities. At the global minima, two possible geometric configurations (gradients of neural conserved quantities) are shown in FIG.~\ref{fig:pt_toy1}. It is easy to check that any other configuration has higher loss than at least one of the two configurations. Which configuration has lower loss depends on $\lambda$: when $\lambda<\frac{1-\epsilon}{2}$, two networks represent the same function (i.e., the only conserved quantity); when $\lambda>\frac{1-\epsilon}{2}$, two networks represent two independent functions, one of which is not a conserved quantity even in the sense of Frobenius integrability. Since only the first phase is desirable, we need to set $\lambda<\frac{1-\epsilon}{2}$. This condition can be easily satisfied if $\epsilon\ll 1$.

{\bf Toy example 2:} The second toy example is inspired by the 2D anisotropic harmonic oscillator. To better visualize the example, we consider a 3D (rather than 4D) phase space, but the intrinsic nature of the problem does not change. There are two conserved quantities in the sense of Frobenius integrability. One is easy for neural networks to fit, hence the approximation error can be minimized to zero; another is hard, so a neural network can at best approximate the function up to an error $\epsilon$. Similarly to the analysis above, three possible configurations are global minima. We train three neural networks to learn the conserved quantities. When $\lambda<\frac{\epsilon}{2}$, three models represent only one conserved quantity (the easy one); when $\frac{\epsilon}{2}<\lambda<1$, three models represent two independent conserved quantities (both the easy and the hard one); when $\lambda>1$, a third false conserved quantity is learned. Both the first phase and the second phase are acceptable, depending on different notions of integrability, since a hard conserved quantity may be locally well-behaved but globally ill-behaved. If we search for globally conserved quantities, the first phase is desired. However, if we allow locally conserved quantities, the second phase is desired. All the experiments in the main text are conducted with $\lambda=0.02$, which is equivalent to saying we only care about conserved quantities whose approximation errors are less than $0.02c$. $c=2$ in the current toy example, but we expect $c\sim O(1)$ in general.

\begin{figure}
    \centering
    \includegraphics[width=1.0\linewidth]{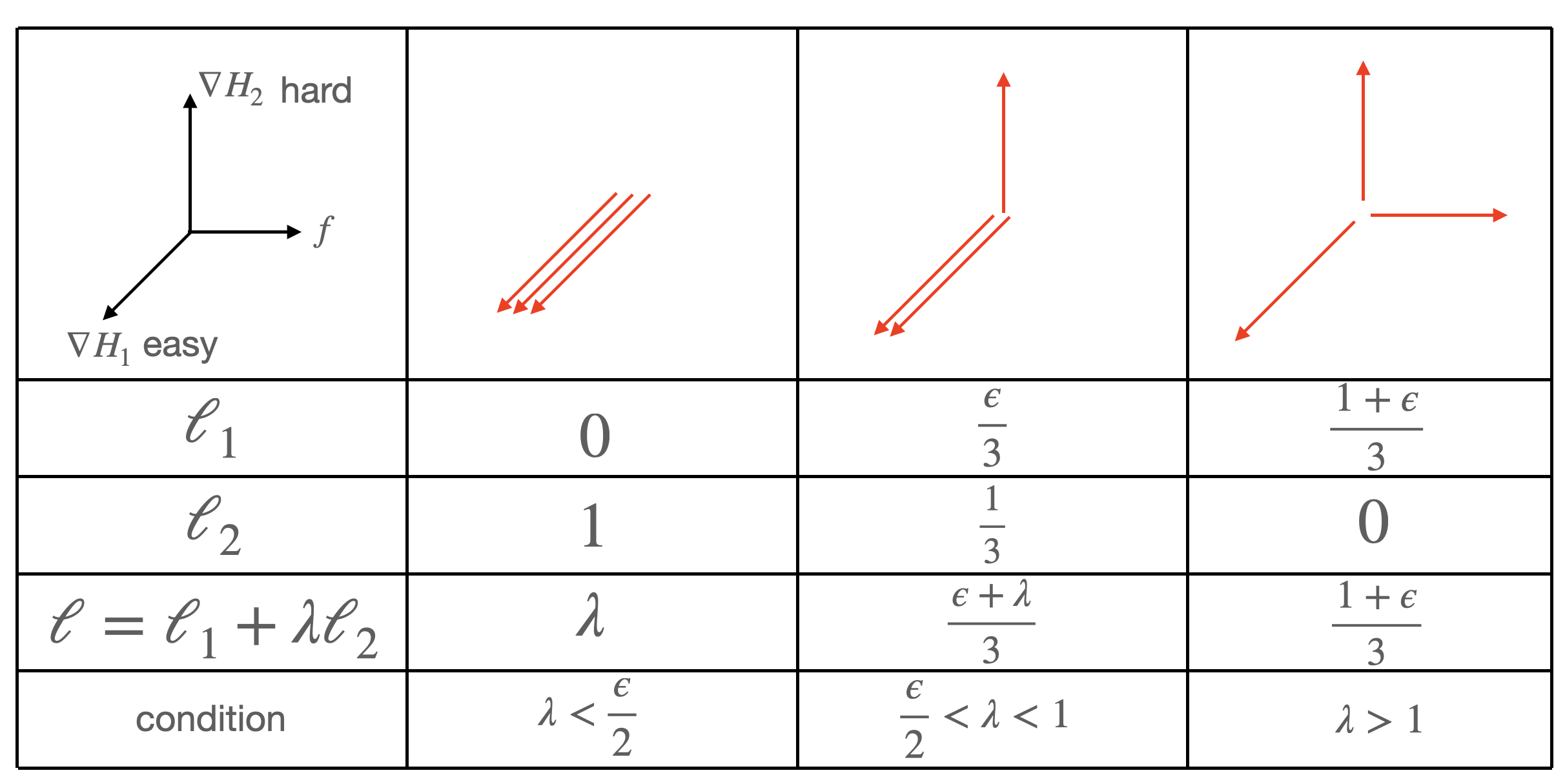}
    \caption{3D Toy example: With different $\lambda$, the global minima may have different geometric configurations. Assume the first and second conserved quantity can be approximated by a neural network with zero error (easy) and $\epsilon>0$ error (hard), respectively.}
    \label{fig:pt_toy2}
\end{figure}

The analysis of two toy examples above suggests a simple picture of phase transitions for more complicated systems: for $n$ conserved quantities with different difficulty (approximation error $\epsilon_1<\epsilon_2<\cdots<\epsilon_n$), we expect there to be $n+1$ phases. At each phase transition, only one conserved quantity is learned or un-learned, and the order of phase transitions depends on the order of $\epsilon$. From the picture of phase transitions, one learns not only the number of conserved quantities, but also knows their difficulty hierarchy. In practice, the phase transition diagram may not be as clean as in these toy examples due to neural network training inefficiency. We show that the phase transition diagram agrees reasonably well with our theory above for the 1D damped harmonic oscillator and 2D harmonic oscillator.  We would like to investigate this further in future work.

\begin{figure}[htbp]
    \centering
    \includegraphics[width=0.8\linewidth]{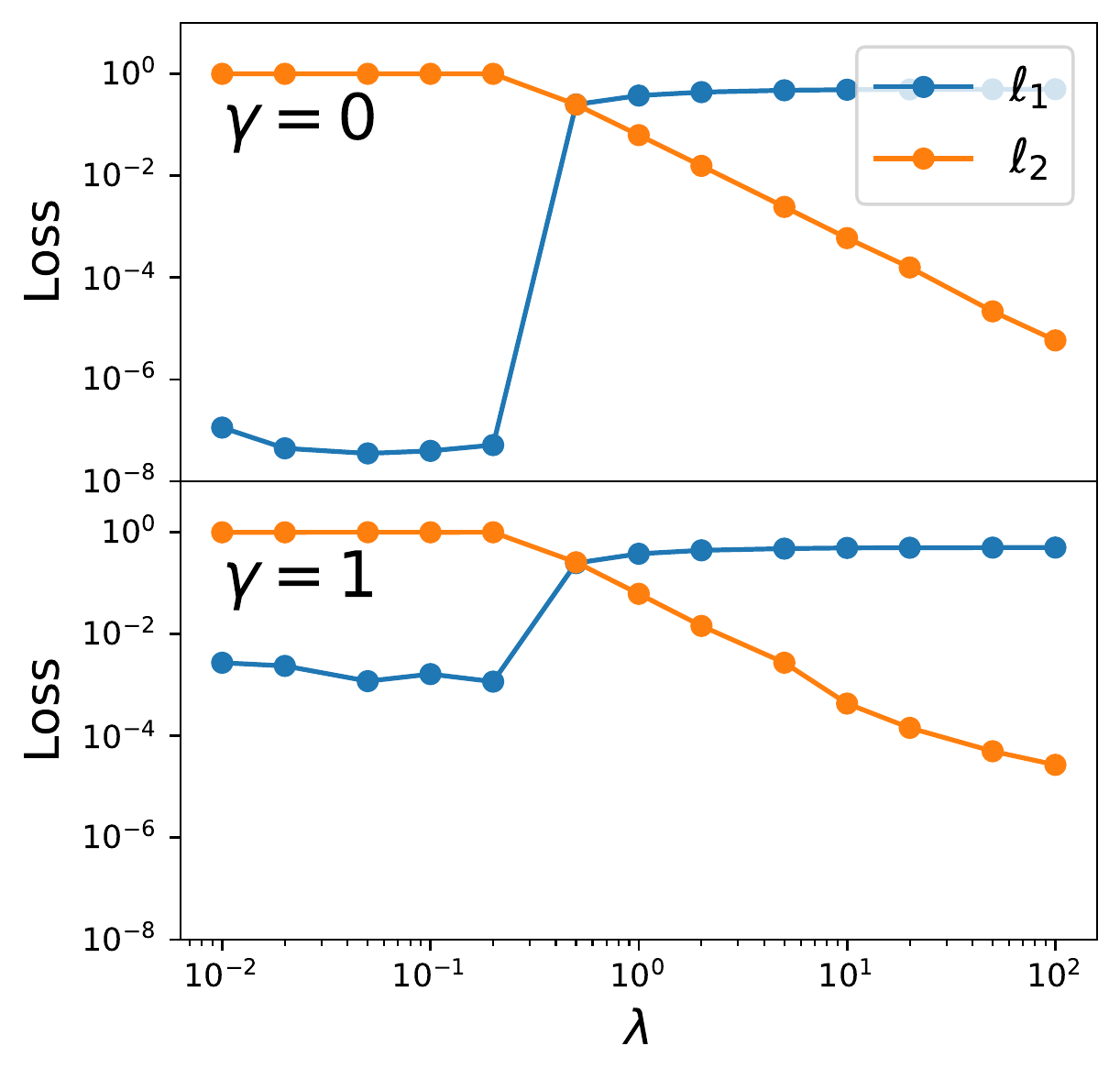}
    \caption{1D damped harmonic oscillator: $\ell_1/\ell_2$ as functions of $\lambda$ demonstrate phase transition behavior.}
    \label{fig:1d_ho_lamb_pt}
\end{figure}

{\bf 1D damped harmonic oscillator}
Toy example 1 can apply to the 1D damped harmonic oscillator without any modification. FIG.~\ref{fig:1d_ho_lamb_pt} shows that we find a phase transition of $\ell_1/\ell_2$ at $\lambda\approx \frac{1}{2}$ for both $\gamma=0$ and $\gamma=1$. When $\gamma=1$, the non-zero $\ell_1$ in the first phase implies the irregularity of the conserved quantity.

\begin{figure}[htbp]
    \centering
    \includegraphics[width=0.8\linewidth]{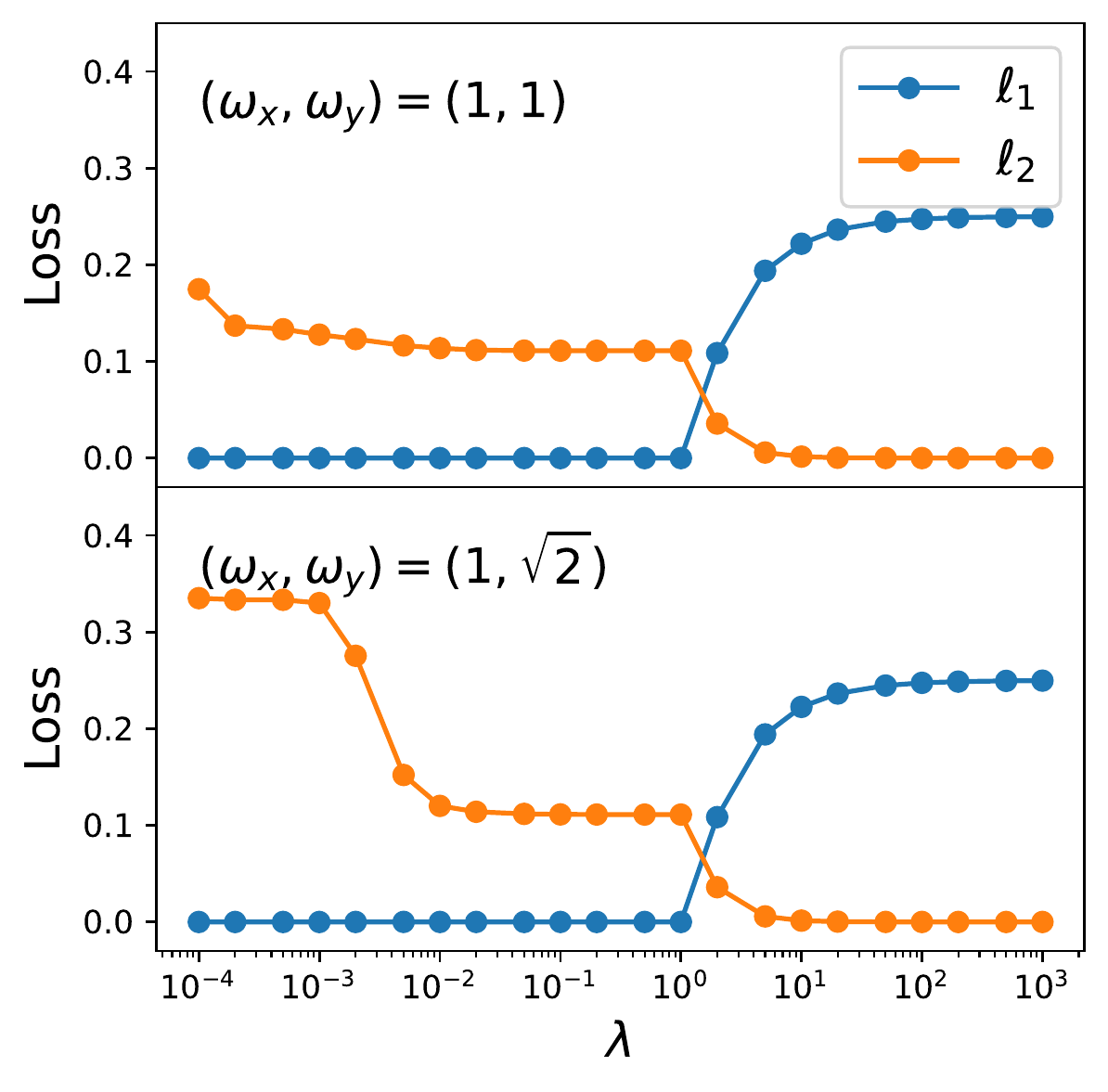}
    \caption{2D isotropic/anisotropic harmonic oscillator: $\ell_1/\ell_2$ as functions of $\lambda$ demonstrate phase transition behavior.}
    \label{fig:2d_ho_lamb_pt}
\end{figure}

{\bf 2D harmonic oscillator}
Toy example 2 is a good abstraction of the 2D harmonic oscillator, but should not be considered to be exact in the quantitative sense. The two energies are easy conserved quantities, while the third conserved quantity regarding phases are harder to learn due to its irregularity when $\omega_y/\omega_x$ is not a fractional number. FIG.~\ref{fig:2d_ho_lamb_pt} shows that: when $(\omega_x,\omega_y)=(1,1)$, only one clear phase transition happens around $\lambda=1$. When $(\omega_x,\omega_y)=(1,\sqrt{2})$, two phase transitions are present, one around $\lambda=1$, another around $10^{-3}<\lambda<10^{-2}$.

\section{Conclusions}
We have presented a method that, given a set of differential equations, can determine not only the number of independent conserved quantities, but also neural (or even symbolic) representations of them. Conservation laws and integrability have many competing definitions listed in Section \ref{app:integrability}, and AI Poincar{\'e} 2.0 is able to adapt to all of them much better than 1.0. In the case of unknown differential equations, however, we have to resort to 1.0. We hope that these tools will may accelerate future progress on exciting open physics problems, for example integrability of quantum many-body systems and many-body localization.

{\bf Acknowledgements} We thank Bohan Wang, Di Luo and Sijing Du for helpful discussions and
the Center for Brains, Minds, and Machines (CBMM)
for hospitality. This work was supported by The Casey
and Family Foundation, the Foundational Questions Institute, the Rothberg Family Fund for Cognitive Science
and IAIFI through NSF grant PHY-2019786.

\bibliography{poincare2.bib}

\begin{thebibliography}{27}%
\makeatletter
\providecommand \@ifxundefined [1]{%
 \@ifx{#1\undefined}
}%
\providecommand \@ifnum [1]{%
 \ifnum #1\expandafter \@firstoftwo
 \else \expandafter \@secondoftwo
 \fi
}%
\providecommand \@ifx [1]{%
 \ifx #1\expandafter \@firstoftwo
 \else \expandafter \@secondoftwo
 \fi
}%
\providecommand \natexlab [1]{#1}%
\providecommand \enquote  [1]{``#1''}%
\providecommand \bibnamefont  [1]{#1}%
\providecommand \bibfnamefont [1]{#1}%
\providecommand \citenamefont [1]{#1}%
\providecommand \href@noop [0]{\@secondoftwo}%
\providecommand \href [0]{\begingroup \@sanitize@url \@href}%
\providecommand \@href[1]{\@@startlink{#1}\@@href}%
\providecommand \@@href[1]{\endgroup#1\@@endlink}%
\providecommand \@sanitize@url [0]{\catcode `\\12\catcode `\$12\catcode
  `\&12\catcode `\#12\catcode `\^12\catcode `\_12\catcode `\%12\relax}%
\providecommand \@@startlink[1]{}%
\providecommand \@@endlink[0]{}%
\providecommand \url  [0]{\begingroup\@sanitize@url \@url }%
\providecommand \@url [1]{\endgroup\@href {#1}{\urlprefix }}%
\providecommand \urlprefix  [0]{URL }%
\providecommand \Eprint [0]{\href }%
\providecommand \doibase [0]{https://doi.org/}%
\providecommand \selectlanguage [0]{\@gobble}%
\providecommand \bibinfo  [0]{\@secondoftwo}%
\providecommand \bibfield  [0]{\@secondoftwo}%
\providecommand \translation [1]{[#1]}%
\providecommand \BibitemOpen [0]{}%
\providecommand \bibitemStop [0]{}%
\providecommand \bibitemNoStop [0]{.\EOS\space}%
\providecommand \EOS [0]{\spacefactor3000\relax}%
\providecommand \BibitemShut  [1]{\csname bibitem#1\endcsname}%
\let\auto@bib@innerbib\@empty
\bibitem [{\citenamefont {Anderson}(1972)}]{PhilipAnderson1972}%
  \BibitemOpen
  \bibfield  {author} {\bibinfo {author} {\bibfnamefont {P.~W.}\ \bibnamefont
  {Anderson}},\ }\bibfield  {title} {\bibinfo {title} {More is different},\
  }\href {https://doi.org/10.1126/science.177.4047.393} {\bibfield  {journal}
  {\bibinfo  {journal} {Science}\ }\textbf {\bibinfo {volume} {177}},\ \bibinfo
  {pages} {393} (\bibinfo {year} {1972})},\ \Eprint
  {https://arxiv.org/abs/https://science.sciencemag.org/content/177/4047/393.full.pdf}
  {https://science.sciencemag.org/content/177/4047/393.full.pdf} \BibitemShut
  {NoStop}%
\bibitem [{\citenamefont {Liu}\ and\ \citenamefont
  {Tegmark}(2021{\natexlab{a}})}]{poincare}%
  \BibitemOpen
  \bibfield  {author} {\bibinfo {author} {\bibfnamefont {Z.}~\bibnamefont
  {Liu}}\ and\ \bibinfo {author} {\bibfnamefont {M.}~\bibnamefont {Tegmark}},\
  }\bibfield  {title} {\bibinfo {title} {Machine learning conservation laws
  from trajectories},\ }\href {https://doi.org/10.1103/PhysRevLett.126.180604}
  {\bibfield  {journal} {\bibinfo  {journal} {Phys. Rev. Lett.}\ }\textbf
  {\bibinfo {volume} {126}},\ \bibinfo {pages} {180604} (\bibinfo {year}
  {2021}{\natexlab{a}})}\BibitemShut {NoStop}%
\bibitem [{\citenamefont {ichi Mototake}(2019)}]{mototake2019interpretable}%
  \BibitemOpen
  \bibfield  {author} {\bibinfo {author} {\bibfnamefont {Y.}~\bibnamefont {ichi
  Mototake}},\ }\bibfield  {title} {\bibinfo {title} {Interpretable
  conservation law estimation by deriving the symmetries of dynamics from
  trained deep neural networks},\ }in\ \href@noop {} {\emph {\bibinfo
  {booktitle} {Machine Learning and the Physical Sciences Workshop at the 33rd
  Conference on Neural Information Processing Systems (NeurIPS)}}}\ (\bibinfo
  {year} {2019})\ \Eprint {https://arxiv.org/abs/2001.00111} {arXiv:2001.00111
  [physics.data-an]} \BibitemShut {NoStop}%
\bibitem [{\citenamefont {Wetzel}\ \emph {et~al.}(2020)\citenamefont {Wetzel},
  \citenamefont {Melko}, \citenamefont {Scott}, \citenamefont {Panju},\ and\
  \citenamefont {Ganesh}}]{PhysRevResearch.2.033499}%
  \BibitemOpen
  \bibfield  {author} {\bibinfo {author} {\bibfnamefont {S.~J.}\ \bibnamefont
  {Wetzel}}, \bibinfo {author} {\bibfnamefont {R.~G.}\ \bibnamefont {Melko}},
  \bibinfo {author} {\bibfnamefont {J.}~\bibnamefont {Scott}}, \bibinfo
  {author} {\bibfnamefont {M.}~\bibnamefont {Panju}},\ and\ \bibinfo {author}
  {\bibfnamefont {V.}~\bibnamefont {Ganesh}},\ }\bibfield  {title} {\bibinfo
  {title} {Discovering symmetry invariants and conserved quantities by
  interpreting siamese neural networks},\ }\href
  {https://doi.org/10.1103/PhysRevResearch.2.033499} {\bibfield  {journal}
  {\bibinfo  {journal} {Phys. Rev. Research}\ }\textbf {\bibinfo {volume}
  {2}},\ \bibinfo {pages} {033499} (\bibinfo {year} {2020})}\BibitemShut
  {NoStop}%
\bibitem [{\citenamefont {Ha}\ and\ \citenamefont
  {Jeong}(2021)}]{ha2021discovering}%
  \BibitemOpen
  \bibfield  {author} {\bibinfo {author} {\bibfnamefont {S.}~\bibnamefont
  {Ha}}\ and\ \bibinfo {author} {\bibfnamefont {H.}~\bibnamefont {Jeong}},\
  }\bibfield  {title} {\bibinfo {title} {Discovering invariants via machine
  learning},\ }\href {https://doi.org/10.1103/PhysRevResearch.3.L042035}
  {\bibfield  {journal} {\bibinfo  {journal} {Phys. Rev. Research}\ }\textbf
  {\bibinfo {volume} {3}},\ \bibinfo {pages} {L042035} (\bibinfo {year}
  {2021})}\BibitemShut {NoStop}%
\bibitem [{\citenamefont {Wang}\ \emph {et~al.}(2019)\citenamefont {Wang},
  \citenamefont {Shen}, \citenamefont {Long},\ and\ \citenamefont
  {Dong}}]{wang2019learning}%
  \BibitemOpen
  \bibfield  {author} {\bibinfo {author} {\bibfnamefont {Y.}~\bibnamefont
  {Wang}}, \bibinfo {author} {\bibfnamefont {Z.}~\bibnamefont {Shen}}, \bibinfo
  {author} {\bibfnamefont {Z.}~\bibnamefont {Long}},\ and\ \bibinfo {author}
  {\bibfnamefont {B.}~\bibnamefont {Dong}},\ }\bibfield  {title} {\bibinfo
  {title} {Learning to discretize: solving 1d scalar conservation laws via deep
  reinforcement learning},\ }\href@noop {} {\bibfield  {journal} {\bibinfo
  {journal} {arXiv preprint arXiv:1905.11079}\ } (\bibinfo {year}
  {2019})}\BibitemShut {NoStop}%
\bibitem [{\citenamefont {Sturm}\ and\ \citenamefont
  {Wexler}(2022)}]{sturm2022conservation}%
  \BibitemOpen
  \bibfield  {author} {\bibinfo {author} {\bibfnamefont {P.~O.}\ \bibnamefont
  {Sturm}}\ and\ \bibinfo {author} {\bibfnamefont {A.~S.}\ \bibnamefont
  {Wexler}},\ }\bibfield  {title} {\bibinfo {title} {Conservation laws in a
  neural network architecture: Enforcing the atom balance of a julia-based
  photochemical model (v0. 2.0)},\ }\href@noop {} {\bibfield  {journal}
  {\bibinfo  {journal} {Geoscientific Model Development}\ }\textbf {\bibinfo
  {volume} {15}},\ \bibinfo {pages} {3417} (\bibinfo {year}
  {2022})}\BibitemShut {NoStop}%
\bibitem [{\citenamefont {Kunin}\ \emph {et~al.}(2020)\citenamefont {Kunin},
  \citenamefont {Sagastuy-Brena}, \citenamefont {Ganguli}, \citenamefont
  {Yamins},\ and\ \citenamefont {Tanaka}}]{kunin2020neural}%
  \BibitemOpen
  \bibfield  {author} {\bibinfo {author} {\bibfnamefont {D.}~\bibnamefont
  {Kunin}}, \bibinfo {author} {\bibfnamefont {J.}~\bibnamefont
  {Sagastuy-Brena}}, \bibinfo {author} {\bibfnamefont {S.}~\bibnamefont
  {Ganguli}}, \bibinfo {author} {\bibfnamefont {D.~L.}\ \bibnamefont
  {Yamins}},\ and\ \bibinfo {author} {\bibfnamefont {H.}~\bibnamefont
  {Tanaka}},\ }\bibfield  {title} {\bibinfo {title} {Neural mechanics: Symmetry
  and broken conservation laws in deep learning dynamics},\ }\href@noop {}
  {\bibfield  {journal} {\bibinfo  {journal} {arXiv preprint arXiv:2012.04728}\
  } (\bibinfo {year} {2020})}\BibitemShut {NoStop}%
\bibitem [{\citenamefont {Goodfellow}\ \emph {et~al.}(2016)\citenamefont
  {Goodfellow}, \citenamefont {Bengio},\ and\ \citenamefont
  {Courville}}]{goodfellow2016deep}%
  \BibitemOpen
  \bibfield  {author} {\bibinfo {author} {\bibfnamefont {I.}~\bibnamefont
  {Goodfellow}}, \bibinfo {author} {\bibfnamefont {Y.}~\bibnamefont {Bengio}},\
  and\ \bibinfo {author} {\bibfnamefont {A.}~\bibnamefont {Courville}},\
  }\href@noop {} {\emph {\bibinfo {title} {Deep learning}}}\ (\bibinfo
  {publisher} {MIT press},\ \bibinfo {year} {2016})\BibitemShut {NoStop}%
\bibitem [{Note1()}]{Note1}%
  \BibitemOpen
  \bibinfo {note} {This seems to imply some `simpler' conservation laws are
  preferred by neural networks over others.}\BibitemShut {Stop}%
\bibitem [{Note2()}]{Note2}%
  \BibitemOpen
  \bibinfo {note} {Although the nonlinear manifold learning method introduced
  in AI Poincar{\'e} 1.0 also applies here, the ways to compute the number of
  conserved quantities $n_c$ is different and actually \protect \textit {dual}.
  In Poincar{\'e} 1.0, $n_c$ is the phase space dimension minus the dimension
  of the trajectory manifold. While in this paper, $n_c$ is equal to the
  dimension of the manifold. Because of this duality, the explained ratio
  diagram (ERD) in Poincar\'{e} 1.0 resembles a hill while in FIG.~\ref
  {fig:examples} the ERD is upside down and resembles a valley.}\BibitemShut
  {Stop}%
\bibitem [{\citenamefont {Udrescu}\ and\ \citenamefont
  {Tegmark}(2020)}]{feynman1}%
  \BibitemOpen
  \bibfield  {author} {\bibinfo {author} {\bibfnamefont {S.-M.}\ \bibnamefont
  {Udrescu}}\ and\ \bibinfo {author} {\bibfnamefont {M.}~\bibnamefont
  {Tegmark}},\ }\bibfield  {title} {\bibinfo {title} {Ai feynman: A
  physics-inspired method for symbolic regression},\ }\href
  {https://doi.org/10.1126/sciadv.aay2631} {\bibfield  {journal} {\bibinfo
  {journal} {Science Advances}\ }\textbf {\bibinfo {volume} {6}},\ \bibinfo
  {pages} {eaay2631} (\bibinfo {year} {2020})},\ \Eprint
  {https://arxiv.org/abs/https://www.science.org/doi/pdf/10.1126/sciadv.aay2631}
  {https://www.science.org/doi/pdf/10.1126/sciadv.aay2631} \BibitemShut
  {NoStop}%
\bibitem [{\citenamefont {Udrescu}\ \emph {et~al.}(2020)\citenamefont
  {Udrescu}, \citenamefont {Tan}, \citenamefont {Feng}, \citenamefont {Neto},
  \citenamefont {Wu},\ and\ \citenamefont {Tegmark}}]{feynman2}%
  \BibitemOpen
  \bibfield  {author} {\bibinfo {author} {\bibfnamefont {S.-M.}\ \bibnamefont
  {Udrescu}}, \bibinfo {author} {\bibfnamefont {A.}~\bibnamefont {Tan}},
  \bibinfo {author} {\bibfnamefont {J.}~\bibnamefont {Feng}}, \bibinfo {author}
  {\bibfnamefont {O.}~\bibnamefont {Neto}}, \bibinfo {author} {\bibfnamefont
  {T.}~\bibnamefont {Wu}},\ and\ \bibinfo {author} {\bibfnamefont
  {M.}~\bibnamefont {Tegmark}},\ }\bibfield  {title} {\bibinfo {title} {Ai
  feynman 2.0: Pareto-optimal symbolic regression exploiting graph
  modularity},\ }\href@noop {} {\bibfield  {journal} {\bibinfo  {journal}
  {Advances in Neural Information Processing Systems}\ }\textbf {\bibinfo
  {volume} {33}},\ \bibinfo {pages} {4860} (\bibinfo {year}
  {2020})}\BibitemShut {NoStop}%
\bibitem [{\citenamefont {Landau}\ and\ \citenamefont
  {Lifshitz}(1976)}]{landau1976mechanics}%
  \BibitemOpen
  \bibfield  {author} {\bibinfo {author} {\bibfnamefont {L.}~\bibnamefont
  {Landau}}\ and\ \bibinfo {author} {\bibfnamefont {E.}~\bibnamefont
  {Lifshitz}},\ }\href@noop {} {\emph {\bibinfo {title} {Mechanics third
  edition}}}\ (\bibinfo {year} {1976})\ Chap.~\bibinfo {chapter}
  {2}\BibitemShut {NoStop}%
\bibitem [{\citenamefont {Kingma}\ and\ \citenamefont
  {Ba}(2014)}]{kingma2014adam}%
  \BibitemOpen
  \bibfield  {author} {\bibinfo {author} {\bibfnamefont {D.~P.}\ \bibnamefont
  {Kingma}}\ and\ \bibinfo {author} {\bibfnamefont {J.}~\bibnamefont {Ba}},\
  }\bibfield  {title} {\bibinfo {title} {Adam: A method for stochastic
  optimization},\ }\href@noop {} {\bibfield  {journal} {\bibinfo  {journal}
  {arXiv preprint arXiv:1412.6980}\ } (\bibinfo {year} {2014})}\BibitemShut
  {NoStop}%
\bibitem [{\citenamefont {Sitzmann}\ \emph {et~al.}(2020)\citenamefont
  {Sitzmann}, \citenamefont {Martel}, \citenamefont {Bergman}, \citenamefont
  {Lindell},\ and\ \citenamefont {Wetzstein}}]{sitzmann2020implicit}%
  \BibitemOpen
  \bibfield  {author} {\bibinfo {author} {\bibfnamefont {V.}~\bibnamefont
  {Sitzmann}}, \bibinfo {author} {\bibfnamefont {J.}~\bibnamefont {Martel}},
  \bibinfo {author} {\bibfnamefont {A.}~\bibnamefont {Bergman}}, \bibinfo
  {author} {\bibfnamefont {D.}~\bibnamefont {Lindell}},\ and\ \bibinfo {author}
  {\bibfnamefont {G.}~\bibnamefont {Wetzstein}},\ }\bibfield  {title} {\bibinfo
  {title} {Implicit neural representations with periodic activation
  functions},\ }\href@noop {} {\bibfield  {journal} {\bibinfo  {journal}
  {Advances in Neural Information Processing Systems}\ }\textbf {\bibinfo
  {volume} {33}},\ \bibinfo {pages} {7462} (\bibinfo {year}
  {2020})}\BibitemShut {NoStop}%
\bibitem [{\citenamefont {Arutyunov}(2019)}]{Arutyunov2019}%
  \BibitemOpen
  \bibfield  {author} {\bibinfo {author} {\bibfnamefont {G.}~\bibnamefont
  {Arutyunov}},\ }\bibinfo {title} {Liouville integrability},\ in\ \href
  {https://doi.org/10.1007/978-3-030-24198-8_1} {\emph {\bibinfo {booktitle}
  {Elements of Classical and Quantum Integrable Systems}}}\ (\bibinfo
  {publisher} {Springer International Publishing},\ \bibinfo {address} {Cham},\
  \bibinfo {year} {2019})\ pp.\ \bibinfo {pages} {1--68}\BibitemShut {NoStop}%
\bibitem [{\citenamefont {Dulock}\ and\ \citenamefont
  {McIntosh}(1965)}]{2dhoiso}%
  \BibitemOpen
  \bibfield  {author} {\bibinfo {author} {\bibfnamefont {V.~A.}\ \bibnamefont
  {Dulock}}\ and\ \bibinfo {author} {\bibfnamefont {H.~V.}\ \bibnamefont
  {McIntosh}},\ }\bibfield  {title} {\bibinfo {title} {On the degeneracy of the
  two-dimensional harmonic oscillator},\ }\href
  {https://doi.org/10.1119/1.1971258} {\bibfield  {journal} {\bibinfo
  {journal} {American Journal of Physics}\ }\textbf {\bibinfo {volume} {33}},\
  \bibinfo {pages} {109} (\bibinfo {year} {1965})},\ \Eprint
  {https://arxiv.org/abs/https://doi.org/10.1119/1.1971258}
  {https://doi.org/10.1119/1.1971258} \BibitemShut {NoStop}%
\bibitem [{\citenamefont {Miura}\ \emph {et~al.}(1968)\citenamefont {Miura},
  \citenamefont {Gardner},\ and\ \citenamefont {Kruskal}}]{KdV_cq}%
  \BibitemOpen
  \bibfield  {author} {\bibinfo {author} {\bibfnamefont {R.~M.}\ \bibnamefont
  {Miura}}, \bibinfo {author} {\bibfnamefont {C.~S.}\ \bibnamefont {Gardner}},\
  and\ \bibinfo {author} {\bibfnamefont {M.~D.}\ \bibnamefont {Kruskal}},\
  }\bibfield  {title} {\bibinfo {title} {Korteweg‐de vries equation and
  generalizations. ii. existence of conservation laws and constants of
  motion},\ }\href {https://doi.org/10.1063/1.1664701} {\bibfield  {journal}
  {\bibinfo  {journal} {Journal of Mathematical Physics}\ }\textbf {\bibinfo
  {volume} {9}},\ \bibinfo {pages} {1204} (\bibinfo {year} {1968})},\ \Eprint
  {https://arxiv.org/abs/https://doi.org/10.1063/1.1664701}
  {https://doi.org/10.1063/1.1664701} \BibitemShut {NoStop}%
\bibitem [{\citenamefont {Barrett}(2013)}]{Barrett2013TitleT}%
  \BibitemOpen
  \bibfield  {author} {\bibinfo {author} {\bibfnamefont {J.}~\bibnamefont
  {Barrett}},\ }\bibfield  {title} {\bibinfo {title} {Title : The local
  conservation laws of the nonlinear schrodinger equation}\ }(\bibinfo {year}
  {2013})\BibitemShut {NoStop}%
\bibitem [{Note3()}]{Note3}%
  \BibitemOpen
  \bibinfo {note} {Informally speaking, an integrable system is a dynamical
  system with sufficiently many conserved quantities.}\BibitemShut {Stop}%
\bibitem [{\citenamefont {{Wikipedia
  contributors}}(2021{\natexlab{a}})}]{enwiki:1058752403}%
  \BibitemOpen
  \bibfield  {author} {\bibinfo {author} {\bibnamefont {{Wikipedia
  contributors}}},\ }\href@noop {} {\bibinfo {title} {Integrable system ---
  {Wikipedia}{,} the free encyclopedia}},\ \bibinfo {howpublished}
  {\url{https://en.wikipedia.org/w/index.php?title=Integrable_system&oldid=1058752403}}
  (\bibinfo {year} {2021}{\natexlab{a}}),\ \bibinfo {note} {[Online; accessed
  5-February-2022]}\BibitemShut {NoStop}%
\bibitem [{\citenamefont {VICKERS}(2001)}]{vickers_2001}%
  \BibitemOpen
  \bibfield  {author} {\bibinfo {author} {\bibfnamefont {J.}~\bibnamefont
  {VICKERS}},\ }\bibfield  {title} {\bibinfo {title} {Integrable systems:
  Twistors, loop groups, and riemann surfaces (oxford graduate texts in
  mathematics 4) by n. j. hitchin, g. b. segal and r. s. ward: 136 pp.,
  £25.00, isbn 0-19-850421-7 (clarendon press, oxford, 1999).},\ }\href
  {https://doi.org/10.1112/blms/33.1.117} {\bibfield  {journal} {\bibinfo
  {journal} {Bulletin of the London Mathematical Society}\ }\textbf {\bibinfo
  {volume} {33}},\ \bibinfo {pages} {116–127} (\bibinfo {year}
  {2001})}\BibitemShut {NoStop}%
\bibitem [{\citenamefont {Liu}\ and\ \citenamefont
  {Tegmark}(2021{\natexlab{b}})}]{liu2021machine}%
  \BibitemOpen
  \bibfield  {author} {\bibinfo {author} {\bibfnamefont {Z.}~\bibnamefont
  {Liu}}\ and\ \bibinfo {author} {\bibfnamefont {M.}~\bibnamefont {Tegmark}},\
  }\bibfield  {title} {\bibinfo {title} {Machine-learning hidden symmetries},\
  }\href@noop {} {\bibfield  {journal} {\bibinfo  {journal} {arXiv preprint
  arXiv:2109.09721}\ } (\bibinfo {year} {2021}{\natexlab{b}})}\BibitemShut
  {NoStop}%
\bibitem [{\citenamefont {{Wikipedia
  contributors}}(2021{\natexlab{b}})}]{enwiki:ft}%
  \BibitemOpen
  \bibfield  {author} {\bibinfo {author} {\bibnamefont {{Wikipedia
  contributors}}},\ }\href@noop {} {\bibinfo {title} {Frobenius theorem
  (differential topology) --- {Wikipedia}{,} the free encyclopedia}},\ \bibinfo
  {howpublished}
  {\url{https://en.wikipedia.org/w/index.php?title=Frobenius_theorem_(differential_topology)&oldid=1049676730}}
  (\bibinfo {year} {2021}{\natexlab{b}}),\ \bibinfo {note} {[Online; accessed
  5-February-2022]}\BibitemShut {NoStop}%
\bibitem [{\citenamefont {Karniadakis}\ \emph {et~al.}(2021)\citenamefont
  {Karniadakis}, \citenamefont {Kevrekidis}, \citenamefont {Lu}, \citenamefont
  {Perdikaris}, \citenamefont {Wang},\ and\ \citenamefont
  {Yang}}]{karniadakis2021physics}%
  \BibitemOpen
  \bibfield  {author} {\bibinfo {author} {\bibfnamefont {G.~E.}\ \bibnamefont
  {Karniadakis}}, \bibinfo {author} {\bibfnamefont {I.~G.}\ \bibnamefont
  {Kevrekidis}}, \bibinfo {author} {\bibfnamefont {L.}~\bibnamefont {Lu}},
  \bibinfo {author} {\bibfnamefont {P.}~\bibnamefont {Perdikaris}}, \bibinfo
  {author} {\bibfnamefont {S.}~\bibnamefont {Wang}},\ and\ \bibinfo {author}
  {\bibfnamefont {L.}~\bibnamefont {Yang}},\ }\bibfield  {title} {\bibinfo
  {title} {Physics-informed machine learning},\ }\href@noop {} {\bibfield
  {journal} {\bibinfo  {journal} {Nature Reviews Physics}\ }\textbf {\bibinfo
  {volume} {3}},\ \bibinfo {pages} {422} (\bibinfo {year} {2021})}\BibitemShut
  {NoStop}%
\bibitem [{\citenamefont {Liu}\ \emph {et~al.}(2021)\citenamefont {Liu},
  \citenamefont {Chen}, \citenamefont {Du},\ and\ \citenamefont
  {Tegmark}}]{liu2021physics}%
  \BibitemOpen
  \bibfield  {author} {\bibinfo {author} {\bibfnamefont {Z.}~\bibnamefont
  {Liu}}, \bibinfo {author} {\bibfnamefont {Y.}~\bibnamefont {Chen}}, \bibinfo
  {author} {\bibfnamefont {Y.}~\bibnamefont {Du}},\ and\ \bibinfo {author}
  {\bibfnamefont {M.}~\bibnamefont {Tegmark}},\ }\bibfield  {title} {\bibinfo
  {title} {Physics-augmented learning: A new paradigm beyond physics-informed
  learning},\ }\href@noop {} {\bibfield  {journal} {\bibinfo  {journal} {arXiv
  preprint arXiv:2109.13901}\ } (\bibinfo {year} {2021})}\BibitemShut {NoStop}%
\end{thebibliography}%

\newpage

\onecolumngrid

\appendix 

\newpage

\section{How to determine (in)dependence of multiple conserved quantities}\label{app:symbolic_dependence}

Suppose we know $n$ independent conserved quantities $\mathcal{H}=\{H_1(\z),\cdots,H_n(\z), z\in\mathbb{R}^s\}$, which are parameterized as neural networks or symbolic formulas. How do we determine whether another conserved quantity $H_{n+1}(\z)$ is dependent on or independent of $\mathcal{H}$?

{\bf Method A: differential rank}. We know that $ k_D(\mathcal{H})=n$ due to the functional independence of $\mathcal{H}$. We then compute $k'\equiv k_D(\mathcal{H}\bigcup H_{n+1})$. If $k'=n+1$, then $H_{n+1}$ is independent of $H_n$; otherwise $k'=n$, and $H_{n+1}$ is dependent on $\mathcal{H}$. In practice, we compute the singular value decomposition of $\mat{B}$ (defined in Eq.~(\ref{eq:B})). If the smallest singular value $\sigma_{n+1}<\epsilon_\sigma=10^{-3}$, we consider it vanishing, implying that $k'=n$; otherwise $k'=n+1$. However the complexity of SVD is $O(sn^2)$, which is more computationally expensive than method B.

{\bf Method B: orthogonality test.} Because $\mathcal{H}$ is an independent set of functions, their gradients at almost all $\z$ should span a linear subspace $\mathcal{S(\z)}\equiv{\rm span}(\nabla H_1(\z),\cdots,\nabla H_n(\z))$ of dimensionality $n$. We construct a random unit vector $\widehat{\mat{t}}(\z)$ that is orthogonal to $\mathcal{S}$, which can be computed via a Gram-Schmidt process of a random vector and $n$ gradient vectors. If $H_{n+1}(\z)$ is not independent of $\mathcal{H}$, then the gradient $\nabla H_{n+1}(\z)\in \mathcal{S}(\z)$, so $\widehat{\mat{t}}\cdot \widehat{\nabla H_{n+1}}(\z)=0$. We consider $H_{n+1}$ to be not independent if $\left|\widehat{\mat{t}}(\z)\cdot \widehat{\nabla H_{n+1}}(\z)\right|<\epsilon_i=10^{-3}$ and reject it. If $H_{n+1}(\z)$ is independent of $\mathcal{H}$, then $\left|\widehat{\mat{t}}(\z)\cdot \widehat{\nabla H_{n+1}}(\z)\right|>\epsilon_i$ is true with high probability. To further reduce probability of errors, one may test on $n_t$ points, which incurs an $O(n_t s)$ computational cost.

Once $H_{n+1}$ is verified as being independent of $\mathcal{H}$, we append $H_{n+1}$ to $\mathcal{H}$. This process is repeated until $\left|\mathcal{H}\right|$ (the number of functions) equals the number of conserved quantities (obtained from the neural network front) or the brute force search reaches its computation limit.

\section{Does overfitting happen?}
\begin{figure}[htbp]
    \centering
    \includegraphics[width=0.5\linewidth]{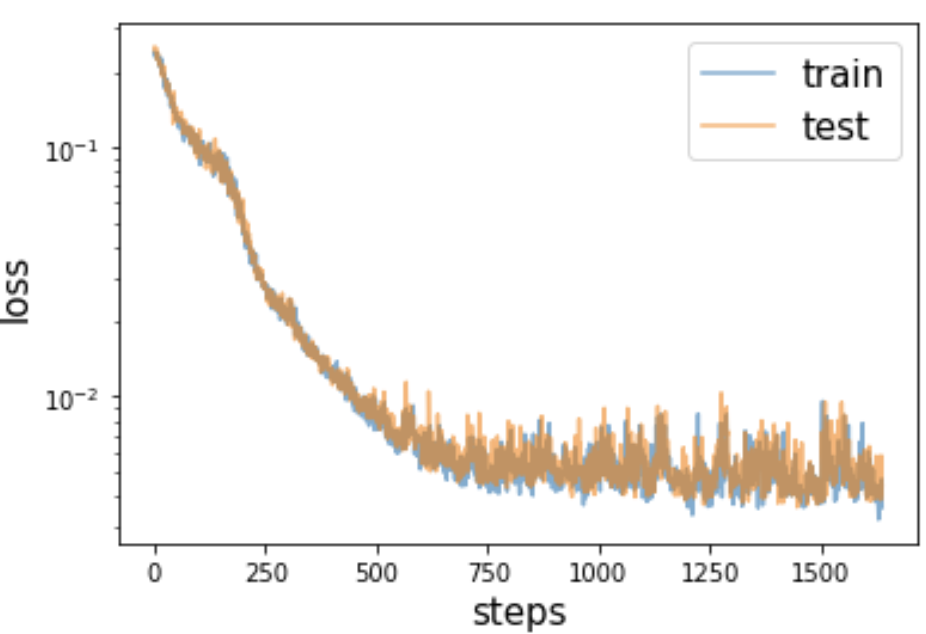}
    \caption{The evolution of the loss function during training, for training data (blue) and testing data (orange). There is no clear generalization gap, implying that overfitting did not happen.}
    \label{fig:threebody_test}
\end{figure}

We split the whole dataset into 50/50 training/testing. FIG.~\ref{fig:threebody_test} shows the result for the three-body problem. Training and testing losses have no gap, signifying that overfitting does not occur.

\end{document}